\documentclass[lettersize,journal]{IEEEtran}
\usepackage{amsmath,amsfonts}
\usepackage{algorithmic}
\usepackage{algorithm}
\usepackage{array}
\usepackage[caption=false,font=normalsize,labelfont=sf,textfont=sf]{subfig}
\usepackage{textcomp}
\usepackage{stfloats}
\usepackage{url}
\usepackage{verbatim}
\usepackage{graphicx}
\usepackage{cite}
\usepackage{bm}
\usepackage{multirow}
\newtheorem{theorem}{\bf Theorem}
\newtheorem{lemma}{\bf Lemma}
\newtheorem{remark}{\bf Remark}

\begin{document}
\graphicspath{{./Figure/}}

\title{Fair Primal Dual Splitting Method for Image Inverse Problems}

\author{Yunfei~Qu,~		Deren~Han%,~\IEEEmembership{Member,~IEEE}
        % <-this % stops a space

\thanks{This research is supported by the National Key R\&D Program of China (No. 2021YFA1003600), the National Natural Science Foundation of China (NSFC) grants  12131004, 12126603, the R\&D project of Pazhou Lab (Huangpu) (Grant no. 2023K0604). (Corresponding author: Deren Han)}
\thanks{Yunfei Qu and Deren Han are with the LMIB, School of Mathematical Sciences, Beihang University, Beijing 100191, China (e-mail: yunfei$\_$math@hotmail.com; handr@buaa.edu.cn).}

\thanks{Manuscript received XX, 2021%; revised August 16, 2021
.}
}

% The paper headers
\markboth{Journal of \LaTeX\ Class Files,~Vol.~14, No.~8, August~2021}%
{Shell \MakeLowercase{\textit{et al.}}: Fair Primal-Dual Splitting Method for Image Processing}

\IEEEpubid{0000--0000/00\$00.00~\copyright~2021 IEEE}
% Remember, if you use this you must call \IEEEpubidadjcol in the second
% column for its text to clear the IEEEpubid mark.

\maketitle

\begin{abstract}
Image inverse problems have numerous applications,  including image processing, super-resolution, and computer vision, which are important areas in image science. These application models can be seen as a three-function composite optimization problem solvable by a variety of primal dual-type methods. We propose a fair primal dual algorithmic framework that incorporates the smooth term not only into the primal subproblem but also into the dual subproblem. We unify the global convergence and establish the convergence rates of our proposed fair primal dual method. Experiments on image denoising and super-resolution reconstruction demonstrate the superiority of the proposed method over the current state-of-the-art.
\end{abstract}

\begin{IEEEkeywords}
Image inverse problem, Splitting method, Fair primal dual, Total variation.
\end{IEEEkeywords}

\section{Introduction}
\IEEEPARstart{I}{mage} inverse problems refer to the task of reconstructing or estimating unknown images from observed measurements obtained through imaging systems. The problem can be mathematically formulated as finding the inverse mapping $\Phi^{-1}$ that satisfies
\begin{equation}\label{inverse model}
	\bm{y = \Phi x+ n},
\end{equation}
where $\bm{y}$ represents the observed measurement, $\bm{x}$ is the unknown original image, $\bm{\Phi}$ represents the operator that maps the image to the measurement, and $\bm{n}$ is the noise. The image inverse problems have a wide variety of applications, including medical imaging, remote sensing, computer vision, and astrophysics.

In general, directly reconstructing the unknown $\bm{x}$ from \eqref{inverse model} is challenging due to the ill-posed nature of the problem, which often leads to an abundance of feasible solutions. To tackle this issue, it is crucial to incorporate a suitable prior or regularizer for $\bm{x}$. Given $\bm{y}$ and $\bm{\Phi}$, recovering $\bm{x}$ from \eqref{inverse model} is equivalent to solving the following unconstrained optimization problem,
\begin{equation}\label{opt problem}
	\min_{\bm{x}} \frac{1}{2}\|\bm{\Phi x-y}\|^2+\lambda \Omega(\bm{x}),
\end{equation}
where $\Omega(\cdot)$ is the regularizer on $\bm{x}$ that incorporates prior knowledge (such as sparsity) of the object $\bm{x}$, and $\lambda>0$ is a trade-off parameter. Among the most important features in images are their edges and texture. Hence, many kinds of regularizer $\Omega(\cdot)$, such as the total variation (TV) norm, wavelet frame-based sparse priors and Gaussian mixed models, have been incorporated in different models in image processing to preserve sharp discontinuities in their solutions in order to keep precise identification of image edges. Below we list three motivated examples.

\IEEEpubidadjcol
 
\begin{itemize}
	\item \emph{Constrained TV model.} The TV regularization, first proposed by Rudin \textit{et al.} \cite{ROF92}, has been applied to a variety of image processing tasks, including image denoising \cite{WYYZ08,ZL11,DZS+11}, inpainting \cite{SC02,JCW+21}, and deconvolution \cite{CW98,PF14,GTS+19}. The constrained TV model can be mathematically formulated as follows
	\begin{equation}\label{CTV}
		\min_{\bm{x}\in\bm{\mathcal{B}}} \frac{1}{2}\|\bm{\Phi x-y}\|^2+\lambda \boldsymbol{TV}(\bm{x}),
	\end{equation}
	where $\bm{\Phi}$ represents the blur operator, mask operator, or convolution matrix; $\boldsymbol{TV}(\cdot)$ represents the isotropic TV norm defined in \eqref{TV_norm}; $\bm{\mathcal{B}=[l,u]}$ is a box area in $\mathbb{R}^n$; and $\lambda$ is a positive trade-off parameter between the data-fidelity and TV regularized term.
	
	\item \emph{Low-Rank Total Variation (LRTV) model.} The LRTV model \cite{SCW+13,SCW+15,ZYY+20} is applied to image super-resolution (SR) reconstruction on medical images, such as CT, MRI, and PET. The LRTV model not only uses the TV norm to utilize information from local neighborhoods but also uses the kernel norm to capture useful information from remote voxels. Mathematically, the LRTV model can be formulated as follows:
	\begin{equation}\label{LRTV}
		\min_{\bm{x}} \frac{1}{2}\|\bm{DSx-T}\|^2 + \lambda_1\|\bm{x}\|_*+\lambda_2\boldsymbol{TV}(\bm{x}),
	\end{equation}
	where $\bm{T}$ denotes the observed lower-resolution image, $\bm{D}$ is a down-sampling operator, $\bm{S}$ is a blurring operator, and $\lambda_1$ and $\lambda_2$ are the respective positive parameters for the low-rank term and TV term.
	
	\item \emph{Weighted Nonlocal Low-Rank Adaptive Total Variation (WNLRATV) model.} The WNLRATV model \cite{CCP+22} is a hyperspectral image (HSI) denoising model that simultaneously considers the spatial-spectral prior and the HSI noise characteristics. Hence, the HSI denoising model can be expressed as follows:
	\begin{equation}\label{WNLRATV}
		\min_{\bm{x}} \|\bm{W}\odot(\bm{x-y})\|_F^2+\lambda_1\|\bm{S}\odot(\bm{\nabla x})\|_1+\lambda_2\|\bm{x}\|_{\text{NSS}},
	\end{equation}
	where $\bm{W}$ is a known tensor, $\bm{S}$ is the weight tensor, $\bm{\nabla}$ is a 3-D first-order forward
	finite-difference operator, $\odot$ represents the element-wise product, $\|\cdot\|_{\text{NSS}}$ is the regularization term to encode the nonlocal spatial similarity and the spatial-spectral correlation, and $\lambda_1$ and $\lambda_2$ are the parameters to control the nonlocal low-rank term and the adaptive TV term, respectively.
\end{itemize}
\IEEEpubidadjcol

In fact, all of the regularization models \eqref{CTV}, \eqref{LRTV}, \eqref{WNLRATV} can be seen as the following composite optimization problem:
\begin{equation}\label{three problem}
	\min_{x} f(x)+h(x)+g(Kx),
\end{equation}
where $f(\cdot)$ is usually a smooth term that has a Lipschitz continuous gradient and serves as a data-fidelity term; $h(\cdot)$ and $g(\cdot)$ represent regularization terms that preserve some special structures (e.g., sharp edges and low-rankness) of solutions; and $K$ is a bounded linear operator.

Recently, the problem \eqref{three problem} has attracted significant attention due to its wide applications. To exploit the three-block structure and smoothness of $f$, some efficient primal dual splitting methods have been proposed to solve problem \eqref{three problem}. Condat \cite{Condat13} and V{\~u} \cite{Vu13} first proposed a primal dual splitting framework (called Condat-V\~u), which can be seen as a generalization of the Chambolle-Pock primal dual splitting method \cite{CP11}, to solve the problem \eqref{three problem}. Subsequently, Davis-Yin three-operator splitting \cite{DY17}, primal dual fixed point (PDFP) \cite{CHZ16}, asymmetric forward-backward-adjoint splitting (AFBA) \cite{LP17}, and a new primal dual three-operator splitting (PD3O) \cite{Yan18} were proposed to solve the problem \eqref{three problem} more efficiently. Indeed, a common feature of the above splitting methods is to first linearize the smooth function $f$ and then merge it into the $h$- or $g$- subproblem.

It should be noted that most saddle-point problems arising from image processing display unbalanced primal and dual subproblems in the sense that one of the subproblems is much easier than the other (e.g., see \cite{CP11,CP16a}). As studied in \cite{HY21}, the authors reshaped the classic augmented Lagrangian method and balanced its subproblems to greatly speed up convergence when solving a linearly
constrained optimization problem that is a special case of problem \eqref{three problem}. And in the most recent work \cite{HWY22}, the authors developed a ``dual stabilization technique'' for saddle-point problems to balance the primal and dual subproblems. They designed an asymmetric primal dual splitting framework that updates the easier subproblem twice to balance the two subproblems. Another work \cite{SGL22} regarded the dual variable as a ``colleague'' rather than a ``follower'' of the primal variables in the Dual Alternating Direction and Multipliers (D-ADMM). They updated the dual variable by solving the dual problem that is more difficult than one step of gradient ascent in conventional ADMM. Additionally, the class of preconditioned primal-dual algorithms, represented by \cite{MCJ+23}, achieves the goal of alleviating subproblem difficulty by selecting special operator matrices for proximal terms (including \cite{HY21} as a special case). This class can also be considered as a balanced technique for solving problem \eqref{three problem}; see \cite{PC11,LXY21, NO23}. As aforementioned, these primal dual splitting methods, including Condat-V\~u, PDFP, AFBA, and PD3O, for problem \eqref{three problem}, put the linearized part of the smooth function $f$ into one of the subproblems, which is unfair and results in unbalanced subproblems. Therefore, a natural idea to balance the subproblems is to simultaneously import partial information from the data-fidelity term $f$ into the two subproblems.

Based on the previous introduction, in this paper, we propose a fair primal dual algorithmic framework (FPD) for problem \eqref{three problem}, where we not only put the information of the data-fidelity term $f$ into the primal subproblem, but the dual subproblem. However, the new dual subproblem involves the conjugate function of the sum of two functions, that generally has no explicit form, leading to the subproblem being difficult to solve. To implement the FPD method, we employ the Fenchel conjugate theory to propose an inexact fair primal dual algorithmic framework (IFPD) that inexactly solves the dual subproblem until some criterion is satisfied. Then we establish the global convergence for the FPD and IFPD methods. Moreover, we derive the $O(1/N)$ convergence rate based on the primal dual gap. %With metric subregularity that is weaker than the strong convexity, we establish the linear convergence rate of FPD and IFPD methods.

This paper is organized as follows. In Section \ref{Prel}, we provide some notations and preliminaries. The FPD algorithmic framework, along with its inexact version, is presented in Section \ref{Algo}. We also establish the convergence analysis and convergence rates for the FPD methods. Finally, two numerical experiments are implemented in Section \ref{Expe}, and conclusions are given in Section \ref{Conc}.

%
%For example, 
%
%s include image deblurring, where A models the blur kernel, and magnetic resonance imaging (MRI), where the linear operator is usually a combination of a Fourier transform and the coil sensitivities.
%
%Classical optimization algorithms from non-linear optimization, such as gradient methods, Newton or quasi-Newton methods, cannot be used ‘out of the box’ since these algorithms require a certain smoothness of the objective function or cannot be applied to large-scale problems – hence the need for specialized algorithms that can exploit the structure of the problems and lead efficiently to good solutions.

\section{Preliminaries}\label{Prel}
In this section, we summarize some notations and basic concepts that will be used in the subsequent analysis.

Throughout this paper, the superscript symbol $^{``\top"}$ represents the transpose of vectors and matrices. Let $\mathbb{R}^n$ be an $n$-dimensional Euclidean space with inner product $\langle\cdot,\cdot\rangle$. The $\ell_1$ norm, the Euclidean norm, and the infinity norm are denoted by $\|\cdot\|_1$, $\|\cdot\|$, and $\|\cdot\|_{\infty}$, respectively. For a given vector $x\in\mathbb{R}^n$, we define the $M$ norm as $\|x\|_{M}=\sqrt{x^{\top}Mx}$, where $M$ is a symmetric and positive definite (or semi-definite) matrix ($M\succ0$ (or $\succeq0$) for short). For a matrix $M$, we denote $\|M\|$ as its matrix 2-norm. The nuclear norm for a matrix $M$ is $\|M\|_*=\sum_{i}\sigma_i(M)$, where $\sigma_i$ denotes the $i$-th singular value of matrix $M$. For a 2-D image $\bm{x}$, its isotropic TV norm is defined as follows:
\begin{equation}\label{TV_norm}
	\boldsymbol{TV}(\bm{x})=\|\bm{\nabla x}\|_{2,1}=\sum_{i=1}^{n}\left\|\bm{[(\nabla_vx)_i,(\nabla_hx)_i]}\right\|_2,
\end{equation}
where $\bm{\nabla=[\nabla_v^{\top},\nabla_h^{\top}]}$, and $\bm{\nabla_v}$ and $\bm{\nabla_h}$ denote the first order difference matrices in the vertical and horizontal directions, respectively. The difference operator for a 3-D image, $\bm{\nabla=[\nabla_v^{\top},\nabla_h^{\top},\nabla_d^{\top}]}$, is a natural extension of the 2-D image, where $\bm{\nabla_d}$ denotes the first-order difference matrix in the depth direction. 

For a convex function $f:\mathbb{R}^n\to\mathbb{R}$, its \emph{subdifferential} at $x$ is the set-valued operator given as follows:
$$\partial f(x)=\{\xi\vert f(y)\ge f(x)+\xi^{\top}(y-x),\forall y\in\operatorname{dom}f\},$$ 
where $\xi$ is called a subgradient of $f$ at $x$, and $\operatorname{dom}f$ is the effective domain of $f$. A convex function $f$ is $L_f$-smooth if its gradient is $L_f$-Lipschitz continuous, i.e., $\exists~L_f>0$, for all $x,y\in\operatorname{dom}f$, 
$$\|\nabla f(x)-\nabla f(y)\|\le L_f\|x-y\|.$$
The Fenchel conjugate function of $f$, denoted by $f^*$, is defined by
$$
f^*(u) = \sup_x\langle x,u\rangle-f(x).
$$
Then we have $y\in\partial f(x)$ if and only if $x\in\partial f^*(y)$.
The proximal operator of $f$, denoted by $\operatorname{Prox}_{\sigma f}(\cdot)$, is given by
$$\operatorname{Prox}_{\sigma f}(a)=\arg\min_{x\in\mathbb{R}^n} f(x)+\frac{1}{2\sigma}\|x-a\|^2,~~a\in\mathbb{R}^n.$$ 
Next, we recall two elementary identities that will be used later.
The first is Moreau identity, which shows that the proximal operator of a continuous convex function $f$ can be calculated from its Fenchel conjugate $f^*$, and vice versa.
\begin{lemma}\cite{Nes18}
	For any $L_f$-smooth convex function $f$ and $x,y\in\operatorname{dom} f$, the following two inequalities hold,
	\begin{align}
	&	f(y)-f(x)-\langle \nabla f(y),x-y\rangle \le\frac{L_f}{2}\|x-y\|^2, \label{Lsmooth1}\\
	&	f(y)-f(x)-\langle \nabla f(y),x-y\rangle \ge\frac{1}{2L_f}\|\nabla f(x)-\nabla f(y)\|^2. \label{Lsmooth2}
	\end{align}
\end{lemma}
\begin{lemma}\cite{BC11}
	For any proper closed convex function $f$ and $\sigma>0, x\in\operatorname{dom}f$, 
	$$
	x = \operatorname{Prox}_{\sigma f}(x) + \sigma \operatorname{Prox}_{\frac{1}{\sigma}f^*}\left(\frac{x}{\sigma}\right).
	$$
\end{lemma}
\begin{lemma}\label{ident2}
	For any vectors $a, b, c, d$ in the finite-dimensional real vector space $\mathcal{Z}$, and the self-adjoint operator $S:\mathcal{Z}\to\mathcal{Z}$, the following identity holds
	$$
	2\langle a-b,S(c-d)\rangle=(\|a-d\|_S^2-\|a-c\|_S^2)+(\|b-c\|_S^2-\|b-d\|_S^2).
	$$
\end{lemma}

Next, we introduce some basic concepts of saddle point theory.
The corresponding dual formulation of the primal problem \eqref{three problem} is given by \cite[Chapter 15]{BC11}
\begin{equation}\label{dual three problem}
	\max_y\,(f+h)^*(-K^{\top}y)+g^*(y).
\end{equation}
The saddle point problem for primal problem \eqref{three problem} and dual problem \eqref{dual three problem} is 
\begin{equation}\label{saddle1 pro}
	\min_{x}\max_y f(x)+h(x)+\langle Kx,y\rangle-g^*(y).
\end{equation} 
Here we present another equivalent form of the saddle point problem \eqref{saddle1 pro} that is useful for the convergence analysis of the proposed algorithmic framework,
\begin{equation}\label{saddle2 pro}
	\min_{x}\max_y \mathcal{L}(x,y):=f_1(x)+h(x)+\langle x,y\rangle-(\tilde g+f_2)^*(y),
\end{equation} 
where $f=f_1+f_2$, both $f_1$ and $f_2$ are convex and $L_{f_1}$-, $L_{f_2}$-smooth functions, respectively, and $\tilde g=g\circ K$. It can be observed that partial information of $f$ is incorporated into the dual problem.

A vector pair $(\hat{x},\hat{y})$ is a saddle point of problem \eqref{saddle2 pro} if and only if 
\begin{equation}\label{saddle}
	\mathcal{L}(\hat{x},y)\le\mathcal{L}(\hat{x},\hat{y})\le\mathcal{L}(x,\hat{y}),\quad \forall~x,y.
\end{equation}
Throughout this paper, for given $x_1, y_1$, we define
\begin{equation*}
	\begin{cases}
		P_{x_1,y_1}(x):=f_1(x)+h(x)-f_1(x_1)-h(x_1)+\langle x-x_1,y_1\rangle,\\
		D_{x_1,y_1}(y):=(\tilde g+f_2)^*(y)-(\tilde g+f_2)^*(y_1)+\langle y-y_1,-x_1\rangle,
	\end{cases}
\end{equation*}
for any $x,y$. Then, we can rewrite the inequalities \eqref{saddle} as follows
\begin{equation}\label{saddlePD}
	\begin{cases}
		P_{\hat{x},\hat{y}}(x)\ge0,\quad\forall x,\\
		D_{\hat{x},\hat{y}}(y)\ge0,\quad\forall y.
	\end{cases}
\end{equation}
And the primal-dual gap is defined as, for any $v$, 
\begin{align*}
	Gap_{\hat{\mathbf{v}}}(\mathbf{v}):=&P_{\hat{x},\hat{y}}(x)+D_{\hat{x},\hat{y}}(y)\\
	=&\Psi(\mathbf{v})-\Psi(
	\hat{\mathbf{v}})+\langle \mathbf{v}-\hat{\mathbf{v}},F(\hat{v})\rangle\ge0,
\end{align*}
where $\mathbf{v}=\begin{pmatrix}
	x\\y
\end{pmatrix}, \Psi(\mathbf{v})=f_1(x)+h(x)+(\tilde g+f_2)^*(y),F(\mathbf{v})=\begin{pmatrix}
	y\\-x
\end{pmatrix}$. When there is no confusion, we will omit the subscript in $P,D,$ and $Gap$. Note that the functions $P(\cdot),~D(\cdot),$ and $Gap(\cdot)$ are all convex.

\section{Algorithm and Convergence Properties} \label{Algo}
%\begin{algorithm}[H]
%\caption{Weighted Tanimoto ELM.}\label{alg:alg1}
%\begin{algorithmic}
%\STATE 
%\STATE {\textsc{TRAIN}}$(\mathbf{X} \mathbf{T})$
%\STATE \hspace{0.5cm}$ \textbf{select randomly } W \subset \mathbf{X}  $
%\STATE \hspace{0.5cm}$ N_\mathbf{t} \gets | \{ i : \mathbf{t}_i = \mathbf{t} \} | $ \textbf{ for } $ \mathbf{t}= -1,+1 $
%\STATE \hspace{0.5cm}$ B_i \gets \sqrt{ \textsc{max}(N_{-1},N_{+1}) / N_{\mathbf{t}_i} } $ \textbf{ for } $ i = 1,...,N $
%\STATE \hspace{0.5cm}$ \hat{\mathbf{H}} \gets  B \cdot (\mathbf{X}^T\textbf{W})/( \mathbb{1}\mathbf{X} + \mathbb{1}\textbf{W} - \mathbf{X}^T\textbf{W} ) $
%\STATE \hspace{0.5cm}$ \beta \gets \left ( I/C + \hat{\mathbf{H}}^T\hat{\mathbf{H}} \right )^{-1}(\hat{\mathbf{H}}^T B\cdot \mathbf{T})  $
%\STATE \hspace{0.5cm}\textbf{return}  $\textbf{W},  \beta $
%\STATE 
%\STATE {\textsc{PREDICT}}$(\mathbf{X} )$
%\STATE \hspace{0.5cm}$ \mathbf{H} \gets  (\mathbf{X}^T\textbf{W} )/( \mathbb{1}\mathbf{X}  + \mathbb{1}\textbf{W}- \mathbf{X}^T\textbf{W}  ) $
%\STATE \hspace{0.5cm}\textbf{return}  $\textsc{sign}( \mathbf{H} \beta )$
%\end{algorithmic}
%\label{alg1}
%\end{algorithm}
In this section, we first review the original primal dual splitting methods for solving the problem \eqref{saddle1 pro} and propose a fair primal dual splitting algorithmic framework in Algorithm \ref{Alg:FPD}. Then, we establish the convergence and convergence rates for the fair primal dual framework. We also propose an exact version of the fair framework and provide the corresponding convergence and convergence rates.

\subsection{ A Fair Primal Dual splitting method}

As far as we know, Condat-V\~u, PDFP, AFBA, and PD3O are the main primal dual methods for solving the saddle point problem \eqref{saddle1 pro}. To demonstrate the common features of these algorithms, we recast these algorithmic schemes as the following unified algorithm framework:
\begin{equation*}
	\begin{cases}
		\hat{x}^{k+1}=\operatorname{Prox}_{\sigma h}\left(x^k-\sigma(K^{\top}y^k+\nabla f({x}^k))\right),\\		
		\bar{x}^{k+1}= \text{Iteration I},\\
		{y}^{k+1}=\operatorname{Prox}_{\tau g^*}\left(y^k+\tau K\bar{x}^{k+1})\right),\\
		x^{k+1} = \text{Iteration II},
	\end{cases}
\end{equation*}
where $\sigma$ and $\tau$ are the stepsizes of the primal and dual subproblems. For the four primal dual algorithms, Iterations I and II are summarized in Table \ref{Tab:iter}.
\begin{table}[h]
	\caption{The Iterations I and II of the four primal dual algorithms}
	\centering
	\begin{tabular}{cl}
		\hline
		& \multicolumn{1}{c}{Iteration I} \\ \hline
		\textbf{Condat-V\~u}   &   $\bar{x}^{k+1}=2\hat{x}^{k+1}-x^k$  \\
		\textbf{PDFP} &   $\bar{x}^{k+1}=\hat{x}^{k+1}$    \\
		\textbf{AFBA} &   $\bar{x}^{k+1}=\hat{x}^{k+1}$  \\
		\textbf{PD3O} &   $\bar{x}^{k+1}=2\hat{x}^{k+1}-x^k +\sigma\left( \nabla f(x^k)-\nabla f(\hat{x}^{k+1})\right)$  \\ \hline
		\vspace{-7pt} &                               \\\hline
		& \multicolumn{1}{c}{Iteration II} \\ \hline
		\textbf{Condat-V\~u}   & $x^{k+1}=\hat{x}^{k+1} $         \\
		\textbf{PDFP} & ${x}^{k+1}=\operatorname{Prox}_{\sigma h}\left(x^k-\sigma(K^{\top}y^{k+1}+\nabla f({x}^k))\right)$ \\
		\textbf{AFBA} & ${x}^{k+1}=\bar{x}^{k+1}-\sigma K^{\top}(y^{k+1}-y^k)$ \\
		\textbf{PD3O} & $x^{k+1}=\hat{x}^{k+1}$      \\ \hline
	\end{tabular}\label{Tab:iter}
\end{table}

 It can be observed from the iteration schemes that these algorithms first linearize the smooth term $f$ and then incorporate it into the $x$- subproblem, which can be seen as a Forward-Backward iteration \cite{CW05}. However, in this case, the two subproblems are unbalanced since the information of the smooth term $f$ is entirely used in one of the subproblems. Hence, in this paper, we propose a fair primal dual algorithmic framework that has two more balanced subproblems for the equivalent saddle point problem \eqref{saddle2 pro}. Concretely, we only linearize a part of the smooth term $f$ and put it into one subproblem, then merge the rest of $f$ into the other subproblem. The details are summarized in Algorithm \ref{Alg:FPD}.

\begin{algorithm}
	\caption{A Fair Primal Dual splitting algorithmic framework (FPD)}
	\begin{algorithmic}[]
		\STATE Initialize $x^0,y^{0}, \tau,\sigma>0$, %and $\sigma\tau\le1-\sigma L_{f_1}$,
		\STATE \textbf{for }{$k>0:$}
		\begin{subequations}
			\begin{align}	
				&\hat{x}^{k+1}=\operatorname{Prox}_{\sigma h}\left(x^k-\sigma(y^k+\nabla f_1({x}^k))\right),\label{FPDx}\\	
				&\bar{x}^{k+1}= \text{Iteration I'},\notag\\
				&{y}^{k+1}=\operatorname{Prox}_{\tau(\tilde{g}+f_2)^*}\left(y^k+\tau\bar{x}^{k+1}\right),\label{FPDy}\\
				&x^{k+1} = \text{Iteration II'}.\notag
			\end{align}
		\end{subequations}
		\STATE \textbf{end for}
	\end{algorithmic}\label{Alg:FPD}
\end{algorithm}
%\begin{remark}
%	It should be remarked that the FPD method is proposed based on the Condat-V\~u primal dual method but different with it even if $f_2=0$. In fact, we can also propose the fair version of PDFP, AFBA, and PD3O, accordingly.
%\end{remark}
\begin{remark}
	The Iterations I' and II' of Condat-V\~u, PDFP, AFBA, PD3O in fair primal dual framework are the same as in Table \ref{Tab:iter} except for the following two steps.
	\begin{table}[h]
		\centering
		\begin{tabular}{cl}
			\hline
			& \multicolumn{1}{c}{Iteration I'} \\ \hline
			\textbf{PD3O} &   $\bar{x}^{k+1}=2\hat{x}^{k+1}-x^k +\sigma\left( \nabla f_1(x^k)-\nabla f_1(\hat{x}^{k+1})\right)$  \\ \hline
			\vspace{-7pt} &                               \\\hline
			& \multicolumn{1}{c}{Iteration II'} \\ \hline
			\textbf{PDFP} & ${x}^{k+1}=\operatorname{Prox}_{\sigma h}\left(x^k-\sigma(y^{k+1}+\nabla f_1({x}^k))\right)$    \\ 
			\textbf{AFBA} & ${x}^{k+1}=\bar{x}^{k+1}-\sigma (y^{k+1}-y^k)$ \\\hline
		\end{tabular}
	\end{table}
\end{remark}
\begin{remark}
	It can be seen that the main difference between the original primal dual framework and the fair primal dual framework is the $y$ subproblem. We can observe from \eqref{FPDy} that the $y$ subproblem is actually a proximal operator of a conjugate function formed by part of the smooth term $f$ and the composite of $g$ and a linear operator $K$. Though the $y$ subproblem in Algorithm \ref{Alg:FPD} is more difficult to solve than that in the original primal dual algorithm framework, the convergence conditions of Algorithm \ref{Alg:FPD} are more relaxed, which will lead to faster convergence and more efficient performance. The convergence conditions for those original algorithms and fair algorithms are listed as follows.
	\begin{table}[h]
		\centering
		\caption{The comparison of convergence conditions between original and fair algorithms}
		\begin{tabular}{lll} 
			\hline
			\textbf{Algorithm} & \multicolumn{1}{c}{\textbf{Original}} & \multicolumn{1}{c}{\textbf{Fair}} \\ \hline
%			 & \multicolumn{2}{c}{\textbf{Condition C1}} \\ \hline
%			\textbf{Condat-V\~u}        &$\sigma\tau\|K\|^2<1-\sigma L_f/2$            &$\sigma\tau<1-\sigma L_{f_1}/2$     \\
%			\textbf{PDFP,AFBA,PD3O}      &$\sigma\tau\|K\|^2<1,~\sigma L_f<2$          &$\sigma\tau<1,~\sigma L_{f_1}<2$  \\\hline
%			& \multicolumn{2}{c}{\textbf{Condition C2}} \\ \hline
			\textbf{Condat-V\~u}      &$\sigma\tau\|K\|^2<1-\sigma L_f$      		   &$\sigma\tau<1-\sigma L_{f_1}$  \\
			\textbf{PDFP,AFBA,PD3O}      &$\sigma\tau\|K\|^2<1,~\sigma L_f<1$          &$\sigma\tau<1,~\sigma L_{f_1}<1$  \\ \hline
		\end{tabular}\label{Tab:condition}
	\end{table}
\end{remark}

To establish the convergence of the FPD framework, we first present the following key inequality.

\begin{theorem}\label{Thm1:FPD}
	For the sequences $\{(x^k,y^k,\hat{x}^k,\bar{x}^k)\}_{k=1}^{\infty}$ generated by Algorithm \ref{Alg:FPD}, we have that, for all $\mathbf{v}$,
	\begin{align}
		&\Psi(\mathbf{u}^{k+1})-\Psi(\mathbf{v})-\langle \mathbf{v}-\mathbf{u}^{k+1}, F(\mathbf{u}^{k+1})\rangle \notag\\
		&\qquad\qquad \le d(\mathbf{v};\mathbf{v}^{k})-d(\mathbf{v};\mathbf{v}^{k+1})-\tilde{d}(\mathbf{v}^{k+1};\mathbf{v}^{k}).\label{Thm1:key}
	\end{align}
\end{theorem}

\begin{IEEEproof}
	See Appendix \ref{APP:thm1}
\end{IEEEproof}

%\begin{remark}
%	We should point out that the above Theorem is based on the iteration schemes of Condat-V\~u algorithm. Since the different iteration schemes of original primal dual algorithms in Table \ref{Tab:iter}, there will be slightly different key inequalities \eqref{Thm1:key} for other fair primal dual algorithms in Theorem \ref{Thm1:FPD}. To improve readability, we put the key inequalities in Appendix \ref{APP:thm1} for fair PDFP, AFBA, and PD3O algorithms.
%\end{remark}

%Based on Theorem \ref{Thm1:FPD}, we prove the global convergence and the convergence rates in terms of the primal-dual gap of Algorithm \ref{Alg:FPD} under the parameters conditions C1 and C2 in Table \ref{Tab:condition}, respectively.
%
%Beforing demonstrating the global convergence, we first prove that for any $\mathbf{v},\mathbf{v}^k$ and $k\ge0$, $d(\mathbf{v};\mathbf{v}^k)>0$ and $d(\mathbf{v}^{k+1};\mathbf{v}^k)>0$ under the step size conditions in Table \ref{Tab:condition} for the four FPD algorithms, respectively.

To demonstrate the convergence of the FPD algorithmic framework, we also need the following results.
\begin{theorem}\label{Thm2:FPD}
	Consider the sequences $\{(x^k,y^k,\hat{x}^k,\bar{x}^k)\}_{k=1}^{\infty}$ generated by Algorithm \ref{Alg:FPD} with parameter conditions in Table \ref{Tab:condition}, for any $\mathbf{v}$ and $\mathbf{v}$, $k\ge0$,
	$$d(\mathbf{v};\mathbf{v}^k)\ge0,~\tilde{d}(\mathbf{v}^{k+1};\mathbf{v}^k)\ge0,	$$
 where the equations hold if and only if $\mathbf{v}=\mathbf{v}^k$ and $\mathbf{v}^{k+1}=\mathbf{v}^k$.
\end{theorem}

\begin{IEEEproof}
	See Appendix \ref{App:thm2}.
\end{IEEEproof}

\begin{theorem}\label{Thm2+:FPD}
	Consider the sequences $\{(x^k,y^k,\hat{x}^k,\bar{x}^k)\}_{k=1}^{\infty}$ generated by Algorithm \ref{Alg:FPD} with parameter conditions in Table \ref{Tab:condition}, there exists a saddle point $\mathbf{v}^*$ of \eqref{saddle2 pro} such that the sequence $\{\mathbf{v}^k\}_{k=0}^{\infty}$ converges to $\mathbf{v}^*$.
\end{theorem}
\begin{IEEEproof}
	See Appendix \ref{App:thm2+}.
\end{IEEEproof}
\begin{theorem}\label{Thm3:FPD}
	Consider the sequence s$\{(x^k,y^k,\hat{x}^k,\bar{x}^k)\}_{k=1}^{\infty}$ generated by Algorithm \ref{Alg:FPD} with parameter conditions in Table \ref{Tab:condition}, we have that
 for all saddle points $\mathbf{v}^*$ of \eqref{saddle2 pro},
    \begin{equation*}
        \Psi(\mathbf{\hat{u}}^N)-\Psi(\mathbf{v}^*)-\langle \mathbf{v}^*-\mathbf{\hat{u}}^N,F(\mathbf{v}^*)\rangle\le\frac{c_0}{N}\|\mathbf{v}^0-\mathbf{v}^*\|^2,
    \end{equation*}
where $\mathbf{\hat{u}}^N=\frac{1}{N}\sum_{k=1}^{N}{\mathbf{u}}^k,~c_0>0$.
	% \begin{enumerate}
	% 	\item [(i)] (Ergodic) for all saddle points $\mathbf{v}^*$,
	% 	\begin{equation*}
	% 		\Psi(\mathbf{\hat{u}}^N)-\Psi(\mathbf{v}^*)-\langle \mathbf{v}^*-\mathbf{\hat{u}}^N,F(\mathbf{v}^*)\rangle\le\frac{c_0}{N}\|\mathbf{v}^0-\mathbf{v}^*\|^2,
	% 	\end{equation*}
	% 	where $\mathbf{\hat{v}}^N=\frac{1}{N}\sum_{k=1}^{N}{\mathbf{u}}^k,~c_0>0$.
	% 	\item[(ii)] (Pointwise)
	% 	\begin{equation*}
	% 		\|\mathbf{v}^{N}-\mathbf{v}^{N-1}\|^2\le\frac{c_1}{N}\|\mathbf{v}^0-\mathbf{v}^*\|^2, ~c_1>0.
	% 	\end{equation*}
	% \end{enumerate}
\end{theorem}
\begin{IEEEproof}
	See Appendix \ref{App:thm3}.	
\end{IEEEproof}

\subsection{An Inexact Fair Primal Dual splitting method}

In general, the conjugate of the sum of two functions does not have a closed-form expression, and the proximal operator of $(\tilde{g}+f_2)^*$ is consequently difficult to compute. Therefore, we propose an inexact algorithm which is implemented in practice. The details are summarized in Algorithm \ref{Alg:IFPD}.
\begin{algorithm}
	\caption{An Inexact Fair Primal Dual algorithm (IFPD)}
	\begin{algorithmic}[!h]
		\STATE Initialize $x^0,y^{0}, \tau,\sigma>0$, and nonnegative summable sequence $\{\epsilon_{k}\}_{k=1}^{\infty}$,
		\STATE\textbf{for }{$k>0:$}
		\begin{subequations}	
			\begin{align}
				&\hat{x}^{k+1}=\operatorname{Prox}_{\sigma h}\left(x^k-\sigma(y^k+\nabla f_1({x}^k))\right),\label{IFPDx}\\	
				&\bar{x}^{k+1} = \text{Iteration I'}, \notag\\
				&\bar{z}^{k+1}=\frac{1}{\tau}y^k+\bar{x}^{k+1},\label{IFPDzbar}\\
				&\text{Compute }z^{k+1},d^{k+1},y^{k+1}, \text{where}\notag\\
				&d^{k+1}\in\partial\left[\tilde{g}(z)+f_2(z)+\frac{\tau}{2}\|z-\bar{z}^{k+1}\|^2\right]\Big\vert_{z=z^{k+1}},\label{IFPDz}\\
				&y^{k+1}=\tau\bar{z}^{k+1}-\tau z^{k+1}+d^{k+1},\label{IFPDy}\\
				&\text{such that} \notag\\
				&\|d^{k+1}\|\le\frac{\epsilon_{k+1}}{\max\{1,\|y^{k+1}\|\}}.\label{IFPDd} \\
				&x^{k+1}=\text{Iteration II'}. \notag
			\end{align}
		\end{subequations}
		\STATE\textbf{end for}
	\end{algorithmic}\label{Alg:IFPD}
\end{algorithm}
\begin{remark}
	The iteration steps Iteration I' and Iteration II' in the IFPD framework are the same as those in the FPD framework. And the parameter conditions to guarantee convergence of algorithms in the IFPD framework are the same as those in the FPD framework listed in Table \ref{Tab:condition} except for a nonnegative summable sequence $\{\epsilon_{k}\}_{k=1}^{\infty}$.
\end{remark}

Next, we give a key inequality that is similar to the key inequality in Theorem \ref{Thm1:FPD}.

\begin{theorem}\label{Thm4:IFPD}
	For the sequences $\{(x^k,y^k,\hat{x}^k,\bar{x}^k,d^k)\}_{k=1}^{\infty}$ generated by Algorithm \ref{Alg:IFPD}, it holds that for all $\mathbf{v}$,
	\begin{align}
		&\Psi(\mathbf{u}^{k+1})-\Psi(\mathbf{v})-\langle v-\mathbf{u}^{k+1}, F(\mathbf{u}^{k+1})\rangle \notag\\
		\le &d(\mathbf{v};\mathbf{v}^{k})-d(\mathbf{v};\mathbf{v}^{k+1})-\tilde{d}(\mathbf{v}^{k+1};\mathbf{v}^{k})\notag\\
		&+\frac{1}{\tau}\langle y-y^{k+1},d^{k+1}\rangle.\label{Thm4:key}
	\end{align}
\end{theorem}

\begin{IEEEproof}
	From the optimality of \eqref{IFPDx} and the Lipschitz continuity of $\nabla f_1$, we can obtain \eqref{conv1_3}. 
	Substituting \eqref{IFPDy} into \eqref{IFPDz}, we obtain that 
	\begin{equation*}
		y^{k+1}\in\partial (\tilde{g}+f_2)(z^{k+1}),
	\end{equation*}
	which implies that
	\begin{equation*}
		z^{k+1}\in\partial (\tilde{g}+f_2)^*(y^{k+1}).
	\end{equation*}
	From \eqref{IFPDy} and \eqref{IFPDzbar}, we conclude that $y^{k+1}$ is the minimizer of the following problem,
	\begin{equation}\label{conv2_1}
		\min_{y\in\mathcal{R}^m} (\tilde{g}+f_2)^*(y)+\frac{1}{2\tau}\|y-\tau\bar{z}^{k+1}-d^{k+1}\|^2.
	\end{equation}
	It follows from \eqref{IFPDzbar} and the optimality of the problem \eqref{conv2_1} that for any $y\in\mathcal{R}^m$,
	\begin{align}
	&(\tilde{g}+f_2)^*(y)-(\tilde{g}+f_2)^*(y^{k+1})-\langle y-y^{k+1},\bar{x}^{k+1}\rangle \notag \\
	\ge&\langle y-y^{k+1}, \frac{1}{\tau}(y^k-y^{k+1}+d^{k+1})\rangle.\label{conv2_2}
	\end{align}
	Since Iterations I' and II' of IFPD are the same as FPD's, we can derive the key inequality \eqref{Thm4:key} based on \eqref{conv2_2} from a similar analysis to the proof of Theorem \ref{Thm1:FPD}.
%	Then summing \eqref{conv1_3} and \eqref{conv2_2} yields that
%	\begin{align}
%		&\Psi(v^{k+1})-\Psi(v)-\langle v-v^{k+1}, F(v^{k+1})\rangle \notag\\
%		\le&\langle v-v^{k+1},Q(v^{k+1}-v^k)\rangle +\frac{L_{h_1}}{2}\|x^k-x^{k+1}\|^2\notag\\
%		&+\frac{1}{\tau}\langle y-y^{k+1},d^{k+1}\rangle.\label{conv2_3}
%	\end{align}
%	Hence, we obtain \eqref{Thm4:key} from Lemma \ref{ident2}.	
\end{IEEEproof}
Similar to Algorithm \ref{Alg:FPD}, the convergence properties of Algorithm \ref{Alg:IFPD} can be established based on Theorem \ref{Thm4:IFPD}.

%\begin{theorem}\label{Thm5:IFPD}
%	Consider the sequences $\{v^k\}_{k=0}^{\infty}$ generated by Algorithm \ref{Alg:IFPD}, there exists a pair of solutions $v^*$ such that the
%	sequence $\{v^k\}_{k=0}^{\infty}$ converges weakly to $v^*$.
%\end{theorem}
%\begin{IEEEproof}
%	See Appendix \ref{App:thm5}.
%\end{IEEEproof}
\begin{theorem}\label{Thm6:IFPD}
	Consider the sequences $\{(x^k,y^k,\hat{x}^k,\bar{x}^k)\}_{k=1}^{\infty}$ generated by Algorithm \ref{Alg:IFPD}, it holds that
 
	\begin{enumerate}
		\item [(i)] (Convergence.) there exists a saddle point $\mathbf{v}^*$ of \eqref{saddle2 pro} such that the sequence $\{\mathbf{v}^k\}_{k=0}^{\infty}$ converges to $\mathbf{v}^*$.
		\item[(ii)] (Convergence rate.)
		for all saddle points $\mathbf{v}^*$ of \eqref{saddle2 pro},
		\begin{equation*}
			\Psi(\mathbf{\hat{u}}^N)-\Psi(\mathbf{v}^*)-\langle \mathbf{v}^*-\mathbf{\hat{u}}^N,F(\mathbf{v}^*)\rangle=O\left(\frac{1}{N}\right),
		\end{equation*}
		where $\mathbf{\hat{u}}^N=\frac{1}{N}\sum_{k=1}^{N}\mathbf{u}^k$.
	\end{enumerate}
\end{theorem}
\begin{IEEEproof}
	See Appendix \ref{App:thm6}.	
\end{IEEEproof}

\section{Experiment}\label{Expe}

In this section, we aim to demonstrate the effectiveness of our proposed fair primal dual algorithmic framework for solving image problems. We apply the Condat-V\~u, PDFP, AFBA, PD3O, and their inexact fair versions, denoted by FCV, FPDFP, FAFBA, FPD3O, to solve the Constrained TV problems and LRTV problems.
All codes were written in {\sc Matlab} R2022a and all numerical experiments were performed on a personal computer with an i5-8265U processor and 8GB memory.

Before presenting the experiment performance, we first clarify some common settings of different algorithms.
\begin{itemize}
	\item[-] \textbf{Subproblem of Algorithm \ref{Alg:IFPD}.} The key to solving the dual subproblem in Algorithm \ref{Alg:IFPD} is determining a subgradient in \eqref{IFPDz} to satisfy the inexact criterion \eqref{IFPDd}. The process of finding a satisfied subgradient in \eqref{IFPDz} is equivalent to inexactly solving the following convex optimization problem:
	\begin{equation}\label{pro_z}
	\min_z~\tilde{g}(z)+f_2(z)+\frac{\tau}{2}\|z-\bar{z}^{k+1}\|^2.
	\end{equation}
	In fact, there are many inner iterative algorithms to solve problem \eqref{pro_z}. If the proximal operator of $\tilde{g}$ is easy to compute, the proximal gradient (also called Forward-Backword) method and its variants are feasible solvers. Since the appearance of the linear operator $K$ in $\tilde{g}$, the proximal operator of $\tilde{g}$ is generally not easy to compute even if the proximal operator of $g$ has a closed-form solution. In such case, Condat-V\~u, PDFP, AFBA, PD3O, and linearized ADMM are standard solvers for solving problem \eqref{pro_z}.
	
	\item[-] \textbf{Evaluation criteria.} We use the signal-to-noise ratio (SNR) and the structural similarity index (SSIM) \cite{WBS+04} to measure the quality of the restored images. The SNR and SSIM are defined as follows:
	$$
	\text{SNR}=10 \log _{10} \frac{\left\|x^*\right\|_F^2}{\left\|x-x^*\right\|_F^2},
	$$
	and
	$$
	\text { SSIM }=\frac{\left(2 \mu_1 \mu_2+c_1\right)\left(2 \sigma_{12}+c_2\right)}{\left(2 \mu_1^2 \mu_2^2+c_1\right)\left(\sigma_1^2+\sigma_2^2+c_2\right)},
	$$
	where $\|\cdot\|_F$ denotes the Frobenius norm, $x^* \in \mathbb{R}^{m \times n}$ is the original image, ${x} \in \mathbb{R}^{m \times n}$ is the restored image, $c_1>0$ and $c_2>0$ are small constants, $\mu_1$ and $\mu_2$ are the mean values of $x^*$ and ${x}$, respectively; $\sigma_1$ and $\sigma_2$ are the variances of $x^*$ and ${x}$, respectively, and $\sigma_{12}$ is the covariance of $x^*$ and ${x}$. SSIM ranges from 0 to 1, where 1 means perfect recovery.
	
	\item[-] \textbf{Stopping criterion.} We use the relative residual of the primal subproblem to construct the stopping criterion which is defined as follows:
	\begin{equation}\label{stop_cri}
		\text{resi}:=\frac{\|x^{k+1}-x^k\|_F}{\|x^k\|_F}\le\epsilon,
	\end{equation}
	where $\epsilon$ is a given constant. 
\end{itemize}

Before examining the performance of fair methods in real-world image problems, we first apply our fair algorithmic framwork to a synthetic problem, the non-negative constrained Lasso, to identify some characteristics of these methods.
\subsection{Non-negative Lasso}

In this part, we compare the primal dual methods, Condat-V\~{u}, PDFP, AFBA, and PD3O with their fair versions for solving the following non-negative Lasso problem \cite{DYS20}, i.e.,
\begin{equation*}
\min \rho\|x\|_1 + \frac{1}{2}\|Ax-b\|^2,\quad \text{s.t.}~~x\ge0,
\end{equation*} 
where matrix $A\in\mathbb{R}^{m\times n}$, $x\in \mathbb{R}^n,~b\in\mathbb{R}^m$. This problem can be reformulated as the following saddle point problem
\begin{align}
    \min_x\max_y&~ \rho\|x\|_1+\frac{\delta}{2}\|Ax-b\|^2+\langle x,y\rangle \notag\\
    &~-(\delta_{\mathbb{R}^n_+}(\cdot)+\frac{1-\delta}{2}\|A\cdot-b\|^2)^*(y), \label{nlasso}
\end{align}
where $\delta_{\mathbb{R}^n_+}(\cdot)$ is the indicator function of the non-negative space $\mathbb{R}^n_+$.

In our experiments, we randomly generate a sparse vector $\hat{x}$, whose $20\%$ components are set to $1$, and the other components are set to $0$. The matrix $A$ is generated from the standard normal distribution and $b$ is obtained by $b=A\hat{x}+0.01\varepsilon$, where $\varepsilon$ is the standard Gaussian noise. The model parameter in \eqref{nlasso} is set to $\rho=0.01$. For subsequent experiments, the stepsizes for the eight primal dual methods are summarized as follows.
 \begin{itemize}
	\item CV ~\quad $\sigma\tau=1/4,~\sigma=(1-\sigma\tau L)/L_f$;
	\item FCV \quad $\sigma\tau=1/4,~\sigma=(1-\sigma\tau)/L_{f_1}$;
	\item PDFP~~ $\sigma=0.9/L_f,\tau=0.9/(L\sigma)$;
	\item FPDFP~ $\sigma=0.9/L_{f_1},\tau=0.9/\sigma$;
    \item PDFP~~ $\sigma=0.9/L_f,\tau=0.9/(L\sigma)$;
	\item FPDFP~ $\sigma=0.9/L_{f_1},\tau=0.9/\sigma$;
	\item PD3O~~ $\sigma=0.9/L_f,\tau=0.9/(L\sigma)$;
	\item FPD3O~ $\sigma=0.9/L_{f_1},\tau=0.9/\sigma$;
%	\item PDFP~~\quad $\sigma=1.9/L_f,\tau=1.9/L$;
%	\item FPDFP \quad $\sigma=1.9/L_{f_1},\tau=1.9$;
%	\item PD3O~~\quad $\sigma=1.9/L_f,\tau=1.9/L$;
%	\item FPD3O \quad $\sigma=1.9/L_{f_1},\tau=1.9$;
\end{itemize}

Before comparing the original primal dual methods and their fair versions, we first investigate some properties of the fair methods. 

It is known that the $y$-subproblem of Algorithm \ref{Alg:IFPD} requires an inner iterative algorithm for its solution. Here we use the proximal gradient method to solve the $y$-subproblem. To explore the effect of the inner iteration number (denoted by `inn') for the fair primal dual methods, we test the four fair methods for $(m,n)=(3000,5000)$ with different fixed inner iteration numbers: inn=$1,5,10,20$, respectively. The stepsizes are set as before, $\delta=0.5$ and stopping parameter $\epsilon=10^{-6}$. We report the relative residual (denoted by `resi'), the function value (denoted by `F\_value'), the required number of outer iterations (denoted by `Iter.') and running time (denoted by `CPU') in Table \ref{Tab:Nlasso_1}.
\begin{table}[htb]
\caption{Results of fair primal dual methods with different inn for problem \eqref{nlasso}.}
\centering
\begin{tabular}{clcccc}
\hline
Algorithm              & \multicolumn{1}{c}{inn} & resi     & F\_value & Iter. & CPU   \\ \hline
\multirow{4}{*}{FCV}   & 1                       & 9.50E-07 & 1.30E+03 & 194   & 3.12  \\
                       & 5                       & 9.70E-07 & 1.30E+03 & 195   & 7.71  \\
                       & 10                      & 9.20E-07 & 1.30E+03 & 195   & 13.76 \\
                       & 20                      & 9.20E-07 & 1.30E+03 & 195   & 23.11 \\ \hline
\multirow{4}{*}{FPDFP} & 1                       & 9.80E-07 & 1.30E+03 & 147   & 2.44  \\
                       & 5                       & 9.20E-07 & 1.30E+03 & 147   & 5.98  \\
                       & 10                      & 9.30E-07 & 1.30E+03 & 147   & 10.59 \\
                       & 20                      & 9.30E-07 & 1.30E+03 & 147   & 17.51 \\ \hline
\multirow{4}{*}{FAFBA} & 1                       & 8.70E-07 & 1.30E+03 & 147   & 2.86  \\
                       & 5                       & 9.00E-07 & 1.30E+03 & 147   & 5.85  \\
                       & 10                      & 9.10E-07 & 1.30E+03 & 147   & 9.81  \\
                       & 20                      & 9.10E-07 & 1.30E+03 & 147   & 18.97 \\ \hline
\multirow{4}{*}{FPD3O} & 1                       & 8.80E-07 & 1.30E+03 & 147   & 3.69  \\
                       & 5                       & 8.90E-07 & 1.30E+03 & 147   & 6.97  \\
                       & 10                      & 9.00E-07 & 1.30E+03 & 147   & 10.56 \\
                       & 20                      & 9.00E-07 & 1.30E+03 & 147   & 18.14 \\ \hline
\end{tabular}\label{Tab:Nlasso_1}
\end{table}

From Table \ref{Tab:Nlasso_1}, we find that the required number of outer iterations to reach the stopping criterion for the four algorithms is almost the same, and the obtained function values are also the same. However, due to the different numbers of inner iterations, the running time increases as the number of inner iterations increases. Hence, we set inn=1 in the following experiments.

\begin{figure}[h]
    \centering
      \includegraphics[width=0.48\linewidth]{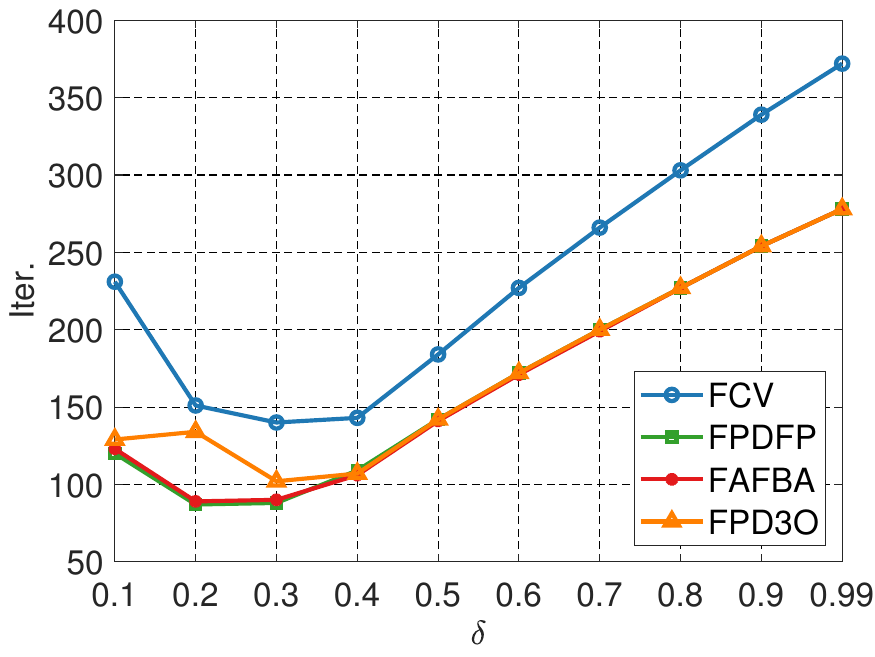}
    \includegraphics[width=0.48\linewidth]{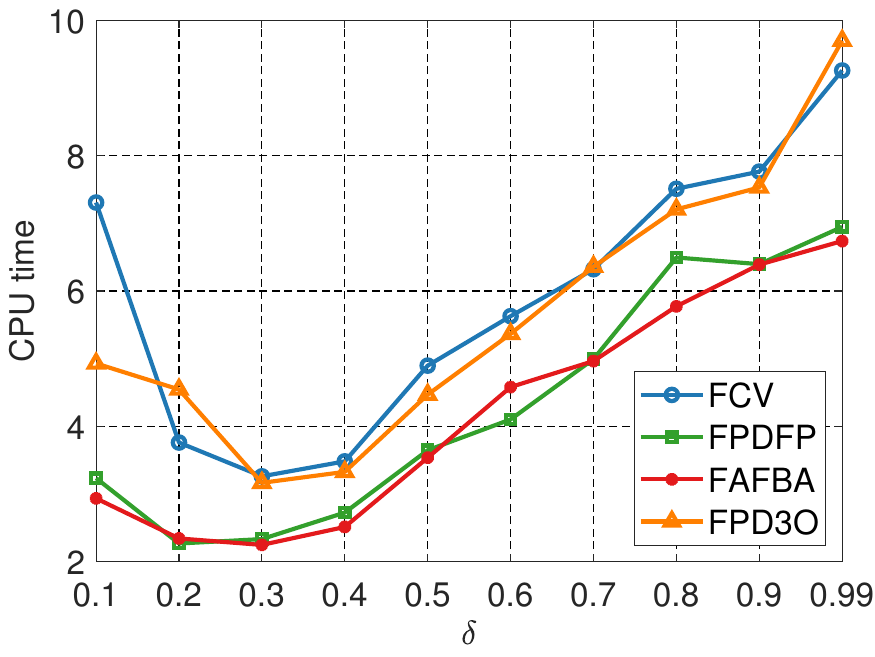}
    \caption{Results of fair primal dual methods with different $\delta$ for problem \eqref{nlasso}}  
    \label{Fig:Nlasso_delta}
\end{figure}
We introduce the information of the smooth term into the two subproblems, and in order to explore the effect of different proportions of information on the performance of the fair algorithms, we compare the numerical effects of the algorithm for different values of $\delta$. We test the four fair methods for $(m,n)=(3000,5000)$ with $\delta=\{0.1,0.2,\cdots,0.9,0.99\}$ and the same stepsizes as before. We set the stopping tolerance $\epsilon=10^{-6}$ and report the required numbers of outer iterations and running time in Fig. \ref{Fig:Nlasso_delta}.

As we can see from Fig. \ref{Fig:Nlasso_delta}, the required numbers of outer iterations and CPU time decrease first and then increase as $\delta$ increases. The optimal values of $\delta$ for the four fair algorithms for solving problem \eqref{nlasso} are between 0.2 and 0.4.

Next, we compare the performance of the original primal dual methods with their fair versions for solving problem \eqref{nlasso}. We test four scenarios for $(m,n)=\{(1000,3000),(3000,5000),(3000,1000),(5000,3000)\}$ with stopping tolerance $\epsilon=10^{-6}$. We set $\delta=0.35$ for the four fair methods, and the stepsizes of the eight primal dual methods are set as before. In Table \ref{Tab:Nlasso_2}, we report the function value, the required number of outer iterations and running time for the eight methods. It can be seen that the fair primal dual methods need fewer iterations and less running
time than their original versions to satisfy the stopping criterion, thus verifying the effectiveness of the fair methods.

\begin{table*}[htb]
\caption{Results of primal dual methods for solving non-negative Lasso problem \eqref{nlasso}}
\centering
\begin{tabular}{lrrrrrrrrrrrrrrr}
\hline
\multicolumn{1}{r}{} & \multicolumn{3}{c}{(1000,3000)}                                       & \multicolumn{1}{c}{} & \multicolumn{3}{c}{(3000,5000)}                                       & \multicolumn{1}{c}{} & \multicolumn{3}{c}{(3000,1000)}                                       & \multicolumn{1}{c}{} & \multicolumn{3}{c}{(5000,3000)}                                       \\ \cline{2-4} \cline{6-8} \cline{10-12} \cline{14-16} 
                     & \multicolumn{1}{c}{F\_value} & \multicolumn{1}{c}{Iter.} & \multicolumn{1}{c}{CPU} & \multicolumn{1}{c}{} & \multicolumn{1}{c}{F\_value} & \multicolumn{1}{c}{Iter.} & \multicolumn{1}{c}{CPU} & \multicolumn{1}{c}{} & \multicolumn{1}{c}{F\_value} & \multicolumn{1}{c}{Iter.} & \multicolumn{1}{c}{CPU} & \multicolumn{1}{c}{} & \multicolumn{1}{c}{F\_value} & \multicolumn{1}{c}{Iter.} & \multicolumn{1}{c}{CPU} \\ \hline
CV                   & 5.2E+02               & 261                   & 0.53                  &                      & 1.3E+03               & 375                   & 4.33                  &                      & 1.0E+03               & 91                    & 0.06                  &                      & 1.8E+03               & 155                   & 1.16                  \\
FCV                  & 5.2E+02               & 110                   & 0.5                   &                      & 1.3E+03               & 143                   & 3.81                  &                      & 1.0E+03               & 55                    & 0.04                  &                      & 1.8E+03               & 72                    & 0.71                  \\
PDFP                 & 5.2E+02               & 195                   & 0.67                  &                      & 1.3E+03               & 281                   & 3.02                  &                      & 1.0E+03               & 68                    & 0.03                  &                      & 1.8E+03               & 116                   & 0.68                  \\
FPDFP                & 5.2E+02               & 69                    & 0.33                  &                      & 1.3E+03               & 92                    & 2.28                  &                      & 1.0E+03               & 34                    & 0.03                  &                      & 1.8E+03               & 43                    & 0.42                  \\
AFBA                 & 5.2E+02               & 195                   & 0.66                  &                      & 1.3E+03               & 281                   & 3.03                  &                      & 1.0E+03               & 68                    & 0.04                  &                      & 1.8E+03               & 116                   & 0.78                  \\
FAFBA                & 5.2E+02               & 73                    & 0.39                  &                      & 1.3E+03               & 93                    & 2.17                  &                      & 1.0E+03               & 34                    & 0.02                  &                      & 1.8E+03               & 44                    & 0.43                  \\
PD3O                 & 5.2E+02               & 195                   & 0.64                  &                      & 1.3E+03               & 281                   & 3.04                  &                      & 1.0E+03               & 68                    & 0.06                  &                      & 1.8E+03               & 116                   & 0.8                   \\
FPD3O                & 5.2E+02               & 75                    & 0.44                  &                      & 1.3E+03               & 104                   & 2.58                  &                      & 1.0E+03               & 38                    & 0.03                  &                      & 1.8E+03               & 47                    & 0.57                  \\ \hline
\end{tabular}\label{Tab:Nlasso_2}
\end{table*}

\begin{table*}[htb]
	\caption{Results of fair primal dual methods with respect to the inner iteration numbers for solving LRTV problem \eqref{LRTV}}
	\centering
	\begin{tabular}{cllcccccccccccccc}
\hline
\multirow{2}{*}{Algorithm} & \multicolumn{1}{c}{\multirow{2}{*}{inn}} &  & \multicolumn{4}{c}{MR1}       & \multicolumn{1}{l}{} & \multicolumn{4}{c}{MR2}       & \multicolumn{1}{l}{} & \multicolumn{4}{c}{MR3}      \\ \cline{4-7} \cline{9-12} \cline{14-17} 
                           & \multicolumn{1}{c}{}                     &  & SNR   & SSIM  & Iter. & CPU   &                      & SNR   & SSIM  & Iter. & CPU   &                      & SNR   & SSIM  & Iter. & CPU  \\ \hline
\multirow{4}{*}{FCV}                        & 1         &  & 22.18 & 0.853 & 61    & 2.91  &                      & 22.97 & 0.801 & 63    & 2.41  &                      & 22.16 & 0.89  & 59    & 2.31 \\
                                            & 5         &  & 22.2  & 0.859 & 61    & 3.82  &                      & 23    & 0.807 & 63    & 3.81  &                      & 22.23 & 0.897 & 60    & 3.65 \\
                                            & 10        &  & 22.19 & 0.859 & 61    & 5.97  &                      & 22.99 & 0.807 & 63    & 5.89  &                      & 22.17 & 0.896 & 59    & 5.65 \\
                                            & 20        &  & 22.34 & 0.86  & 64    & 10.12 &                      & 23.18 & 0.808 & 66    & 10.56 &                      & 22.13 & 0.896 & 58    & 9.49 \\
\multirow{4}{*}{FPDFP}                      & 1         &  & 23.69 & 0.87  & 44    & 2.67  &                      & 24.95 & 0.817 & 46    & 2.18  &                      & 23.45 & 0.9   & 42    & 1.96 \\
                                            & 5         &  & 23.68 & 0.87  & 44    & 3.23  &                      & 24.95 & 0.817 & 46    & 3.34  &                      & 23.38 & 0.9   & 41    & 2.96 \\
                                            & 10        &  & 23.6  & 0.869 & 43    & 4.23  &                      & 24.94 & 0.817 & 46    & 4.37  &                      & 23.38 & 0.899 & 41    & 4.09 \\
                                            & 20        &  & 23.6  & 0.868 & 43    & 7.78  &                      & 24.94 & 0.817 & 46    & 7.22  &                      & 23.44 & 0.898 & 42    & 7.08 \\
\multirow{4}{*}{FPD3O}                      & 1         &  & 24.96 & 0.878 & 33    & 1.54  &                      & 26.35 & 0.833 & 33    & 1.31  &                      & 24.59 & 0.909 & 32    & 1.34 \\
                                            & 5         &  & 24.95 & 0.878 & 33    & 2.39  &                      & 26.33 & 0.833 & 33    & 2.22  &                      & 24.58 & 0.909 & 32    & 2.13 \\
                                            & 10        &  & 24.95 & 0.878 & 33    & 3.63  &                      & 26.33 & 0.833 & 33    & 3.73  &                      & 24.58 & 0.909 & 32    & 3.62 \\
                                            & 20        &  & 24.95 & 0.878 & 33    & 5.44  &                      & 26.33 & 0.833 & 33    & 5.76  &                      & 24.58 & 0.909 & 32    & 5.57 \\ \hline
\end{tabular}
	\label{Tab:inn}
\end{table*}

\subsection{LRTV Problem}

\begin{figure}[htb]
	\centering
\begin{minipage}{0.3\linewidth}
	\centerline{\includegraphics[width=1\textwidth]{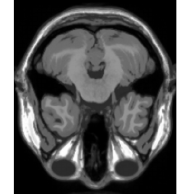}}
	\centerline{MR1}
\end{minipage}
\begin{minipage}{0.3\linewidth}
	\centerline{\includegraphics[width=1\textwidth]{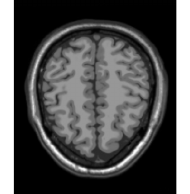}}
	\centerline{MR2}
\end{minipage}
\begin{minipage}{0.3\linewidth}
	\centerline{\includegraphics[width=1\textwidth]{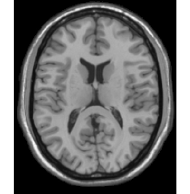}}
	\centerline{MR3}
\end{minipage}
	\caption{The original high-resolution images.}
	\label{Fig:ori}
\end{figure}

We employ the T1 MR phantom from Brainweb to evaluate the recovery performance of the proposed fair methods. Three 217$\times$181 images with a resolution of 1 mm are selected as the test images, as shown in Figure \ref{Fig:ori}. For our experiment, we generate images with 2 mm resolution using the LR image simulation pipeline \cite{SCW+15}, and upsample them again to 1 mm isotropic resolution using the six primal dual methods, respectively. Here we set $\lambda_1=0.01$ and $\lambda_2=0.01$ in LRTV model \eqref{LRTV}.

In this experiment, we compare the original Condat-V\~{u}, PDFP, PD3O, and their fair versions in Algorithm \ref{Alg:IFPD}. We set $f_1(\bm{x})=\frac{\delta}{2}\|\bm{DSx-T}\|^2$ and $f_2(\bm{x})=\frac{(1-\delta)}{2}\|\bm{DSx-T}\|^2$. 
Since we do not know the exact operator norm values of the down-sampled operator $\bm{D}$ and the blurring operator $\bm{S}$, we attempt to ensure the convergence of the proposed iterative algorithm by adjusting $L_f$ as much as possible and finally take $L_f=20$. The operator norm of the discrete gradient norm $L$ equals 8. We tune the parameter $\delta$ and set $\delta=0.9$ for all fair algorithms.
For subsequent experiments, the stepsizes in the six primal dual algorithms are summarized as follows.

\begin{itemize}
	\item CV ~\quad $\sigma\tau=1/80,~\sigma=(1-\sigma\tau L)/L_f$;
	\item FCV \quad $\sigma\tau=1/10,~\sigma=(1-\sigma\tau)/L_{f_1}$;
	\item PDFP~~ $\sigma=0.9/L_f,\tau=0.9/(L\sigma)$;
	\item FPDFP~ $\sigma=0.9/L_{f_1},\tau=0.9/\sigma$;
	\item PD3O~~ $\sigma=0.9/L_f,\tau=0.9/(L\sigma)$;
	\item FPD3O~ $\sigma=0.9/L_{f_1},\tau=0.9/\sigma$;
%	\item PDFP~~\quad $\sigma=1.9/L_f,\tau=1.9/L$;
%	\item FPDFP \quad $\sigma=1.9/L_{f_1},\tau=1.9$;
%	\item PD3O~~\quad $\sigma=1.9/L_f,\tau=1.9/L$;
%	\item FPD3O \quad $\sigma=1.9/L_{f_1},\tau=1.9$;
\end{itemize}
%\begin{table}[htb]
%	\caption{Step sizes setting of the six algorithms}
%	\centering
%	\begin{tabular}{lll}
%		\hline
%		\multicolumn{1}{c}{Algorithm} &  & \multicolumn{1}{c}{Parameter setting}            \\ \hline
%		CV                            &  & $\sigma\tau=1/80,~\sigma=(1-\sigma\tau L)/L_f$   \\
%		FCV                           &  & $\sigma\tau=1/10,~\sigma=(1-\sigma\tau)/L_{f_1}$ \\
%		PDFP                          &  & $\sigma=1.9/L_f,\tau=1.9/L$                      \\
%		FPDFP                         &  & $\sigma=1.9/L_{f_1},\tau=1.9$                    \\
%		PD3O                          &  & $\sigma=1.9/L_f,\tau=1.9/L$                      \\
%		FPD3O                         &  & $\sigma=1.9/L_{f_1},\tau=1.9$                    \\ \hline
%	\end{tabular}\label{Tab:para1}
%\end{table}

For the three fair primal dual methods, FCV, FPDFP, and FPD3O, we employ the Condat-V\~u method to solve the subproblem \eqref{pro_z} inexactly. 
To explore the effect of the number of inner iterations for the fair primal dual methods, we test the three fair methods with different numbers of inner iteration: $\text{inn}=1,5, 10, 20$, respectively. The other stepsize parameters are set as before and $\epsilon=10^{-3}$. The numerical performances, including the final SNR, SSIM, the required number of outer iterations, and computation time are reported in Table \ref{Tab:inn}.

From the results presented in Table \ref{Tab:inn}, we can observe that the image quality recovered by running 1 inner iteration is about the same as running more inner iterations for the three fair primal dual methods. However, the required computation time of each algorithm increases with the number of inner iterations, since the required numbers of outer iterations are almost the same. Hence, we choose $\text{inn}=1$ for fair primal dual methods in the following experiments.

\begin{figure*}[!htb]
	\centering
	\includegraphics[width=0.3\textwidth]{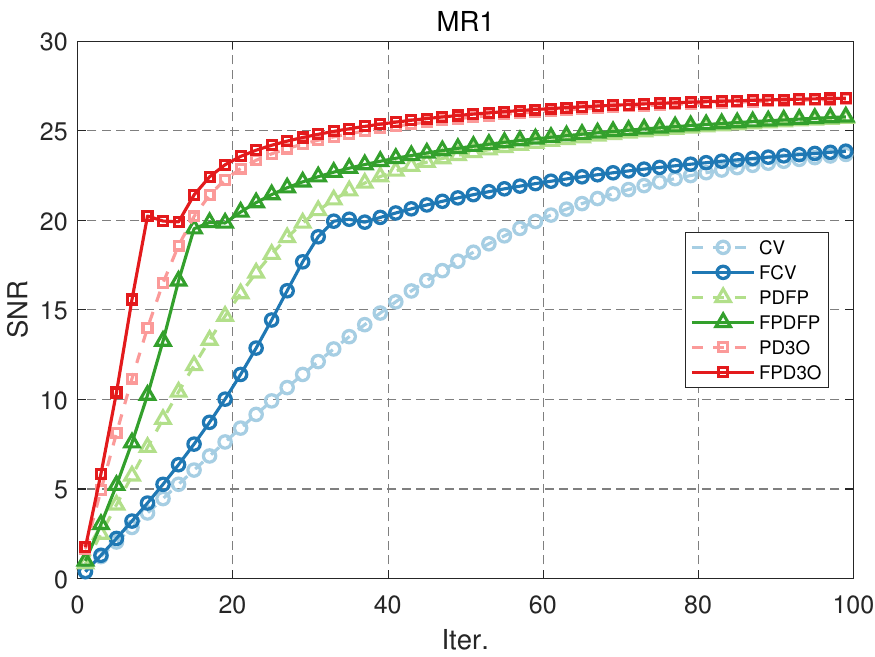}
	\includegraphics[width=0.3\textwidth]{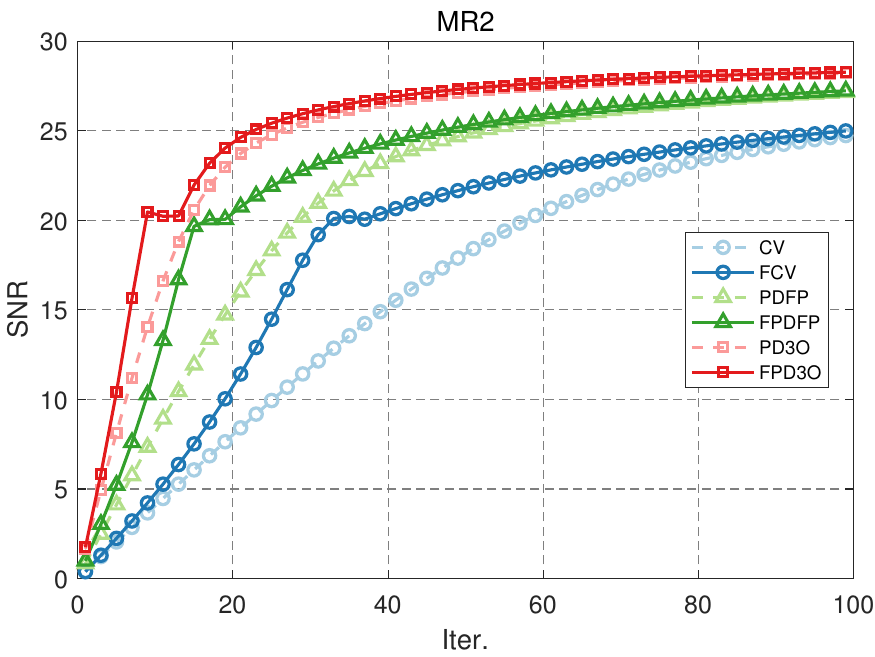}
	\includegraphics[width=0.3\textwidth]{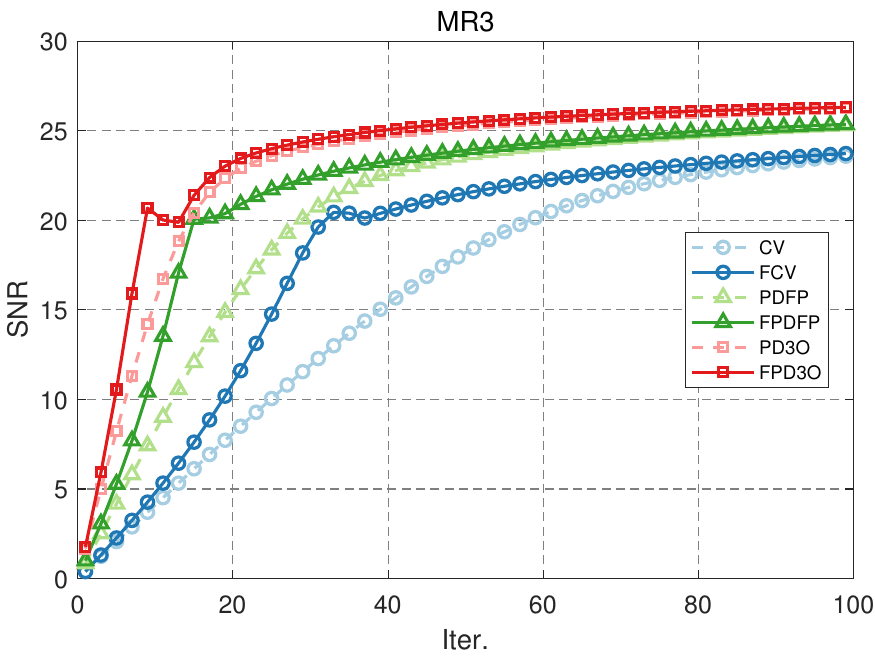}\\
	\includegraphics[width=0.3\textwidth]{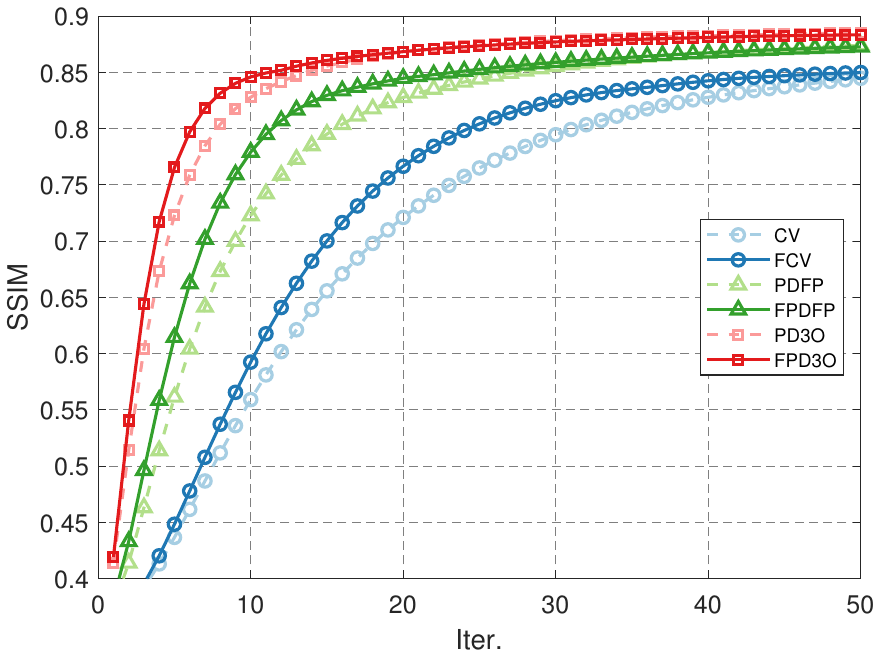}
	\includegraphics[width=0.3\textwidth]{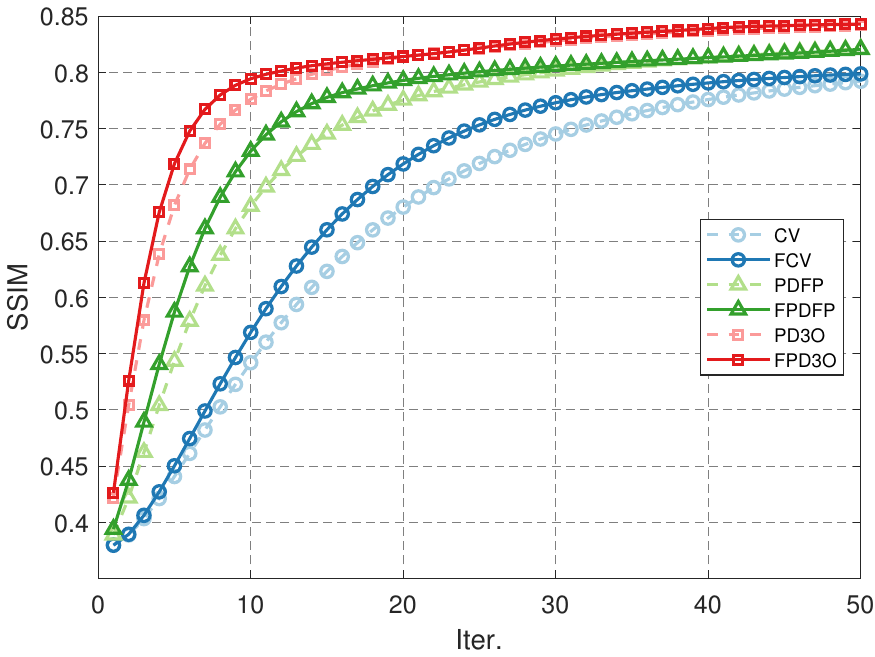}
	\includegraphics[width=0.3\textwidth]{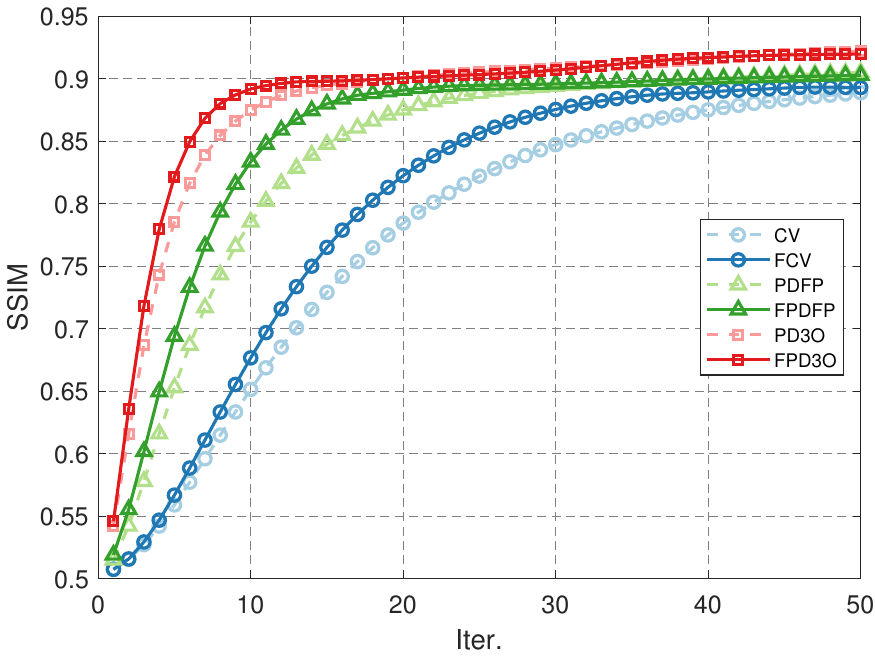}
	\caption{Evolution of SNR and SSIM with respect to iteration numbers for solving LRTV problem.}	
	\label{Fig:SNR_SSIM}
\end{figure*}

To verify the efficiency of our proposed fair primal dual methods, we test the three original methods and the corresponding fair methods on the three images in Fig. \ref{Fig:ori}. In Fig. \ref{Fig:SNR_SSIM}, we graphically show the evolutions of SNR and SSIM values with respect to the iteration numbers. It is clear from Fig. \ref{Fig:SNR_SSIM} that the fair primal dual methods outperform the corresponding original methods in terms of taking fewer iterations to recover images with higher quality.

\subsection{Constrained TV Problem}
In this subsection, we further apply the proposed fair primal dual methods to solve the constrained TV image inpainting problems and compare their numerical performance with the original primal dual methods. Image inpainting refers to the process of filling in missing or damaged regions in images, and it plays a vital role in many image processing tasks \cite{HY12,JCW+21}. The image inpainting problem with TV regularization falls into the model \eqref{CTV} with $\bm{\Phi}$ being the mask operator, i.e., a diagonal matrix whose zero entries denote missed pixels, and identity entries indicate observed pixels. In this case, $\|\bm{\Phi}\|=1$.

\begin{figure}[!h]
	\centering
	\includegraphics[width=0.4\linewidth]{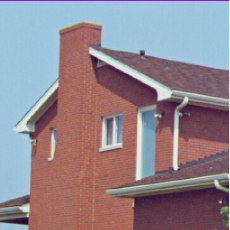}
	\includegraphics[width=0.4\linewidth]{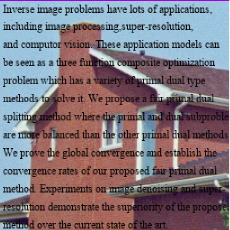}
%	\vspace{-1em}
%	\subfloat{\includegraphics[width=0.4\linewidth]{Fruits0.pdf}}
%	\subfloat{\includegraphics[width=0.4\linewidth]{Fruits1.pdf}}
	\caption{Original House.png (256 $\times$ 256), degraded House.}
	\label{Fig:TV_ori}
\end{figure}

We conducted experiments using the image "House.png" with dimensions 256$\times$256. Following the degradation procedure outlined in \cite{HY12}, the image was first subjected to a masking operation using the operator $\bm{\Phi}$, which simulates the loss of approximately 15\% of the pixels. The image was further degraded by adding zero-mean Gaussian noise with a standard deviation of 0.02. The original and degraded images are shown in Fig. \ref{Fig:TV_ori}.

We specify the parameter selections for the model \eqref{CTV} and tested methods. We set $f_1(\bm{x})=\frac{\delta}{2}\|\bm{\Phi x-y}\|^2$ and $f_2(\bm{x})=\frac{(1-\delta)}{2}\|\bm{\Phi x-y}\|^2$. In this case, $L_f=1$ and $L=8$. We set $\lambda = 0.001$ in the constrained TV model \eqref{CTV}, and set $\delta=0.8$. The other stepsizes for the tested methods are as follows:
\begin{itemize}
	\item AFBA  \quad~$\sigma=0.9/L_f,~\tau=0.9/(L\sigma)$;
	\item FAFBA \quad $\sigma=0.9/L_{f_1},~\tau=0.9/\sigma$;
	\item PDFP  \quad~$\sigma=0.9/L_f,~\tau=0.9/(L\sigma)$;
	\item FPDFP \quad $\sigma=0.9/L_{f_1},\tau=0.9$;
	\item PD3O  \quad~$\sigma=0.9/L_f,~\tau=0.9/(L\sigma)$;
	\item FPD3O \quad $\sigma=0.9/L_{f_1},\tau=0.9$. 
\end{itemize}
Regarding the inner iteration of the fair primal dual methods, we set $\text{inn}=1$ as in the LRTV problem. We continue to use the stopping criterion \eqref{stop_cri} with $\epsilon$ being $10^{-4}$ or $10^{-6}$ in this subsection. We report the overall numerical performance of each tested primal dual algorithm for the constrained TV problem in Table \ref{Tab:CTV}. According to this table, we find out that when $\epsilon=10^{-4}$ in \eqref{stop_cri}, the fair primal dual methods can restore the degraded image with higher SNR and SSIM with similar outer iterations. Furthermore, the fair primal dual methods require fewer iterations to satisfy the stopping criterion \eqref{stop_cri} with $\epsilon=10^{-6}$ and obtain the restored image that is not worse than that of the original primal dual algorithm.
\begin{table}[htb]
	\caption{Results of primal dual methods for constrained TV problem \eqref{CTV}}
	\centering
	\begin{tabular}{llccccccc}
		\hline
		\multirow{2}{*}{} & \multirow{2}{*}{} & \multicolumn{3}{c}{$\epsilon=10^{-4}$}                                                   & \multicolumn{1}{l}{} & \multicolumn{3}{c}{$\epsilon=10^{-6}$}                                                   \\ \cline{3-5} \cline{7-9} 
		&                   & \multicolumn{1}{l}{Iter.} & \multicolumn{1}{l}{SNR} & \multicolumn{1}{l}{SSIM} & \multicolumn{1}{l}{} & \multicolumn{1}{l}{Iter.} & \multicolumn{1}{l}{SNR} & \multicolumn{1}{l}{SSIM} \\ \hline
		\textbf{AFBA}     &                   & 46                        & 20.29                   & 0.89                     &                      & 91                        & 25.63                   & 0.95                     \\
		\textbf{FAFBA}    &                   & 45                        & 23.18                   & 0.93                     &                      & 79                        & 26.63                   & 0.96                     \\
		\textbf{PDFP}     &                   & 46                        & 20.29                   & 0.89                     &                      & 91                        & 25.63                   & 0.95                     \\
		\textbf{FPDFP}    &                   & 45                        & 23.54                   & 0.94                     &                      & 77                        & 26.62                   & 0.96                     \\
		\textbf{PD3O}     &                   & 46                        & 20.19                   & 0.89                     &                      & 91                        & 25.32                   & 0.95                     \\
		\textbf{FPD3O}    &                   & 44                        & 21.21                   & 0.92                     &                      & 87                        & 26.07                   & 0.96                     \\ \hline
	\end{tabular}\label{Tab:CTV}
\end{table}

\begin{figure*}[htb]
	\centering
	\includegraphics[width=0.45\linewidth]{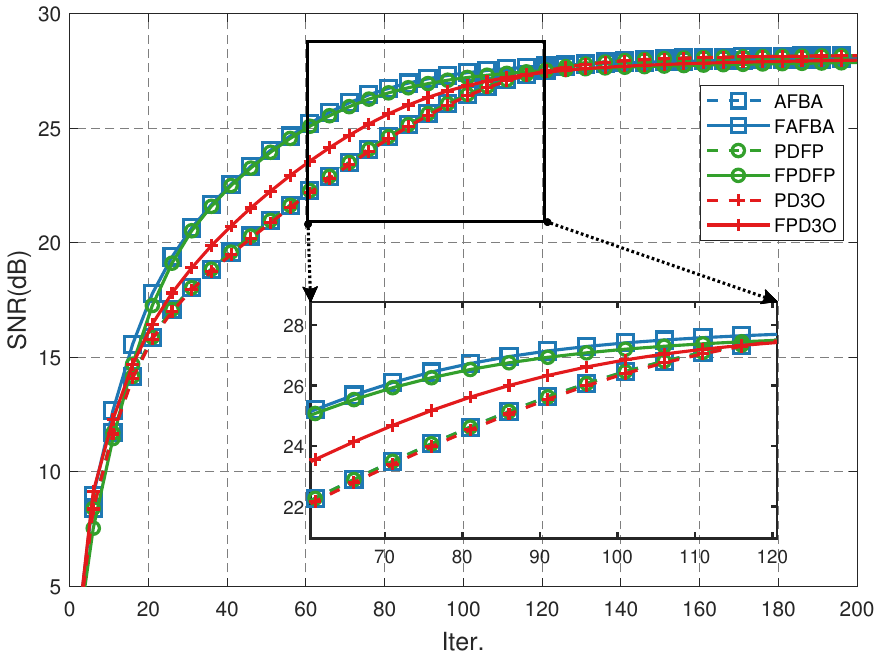}
	\includegraphics[width=0.45\linewidth]{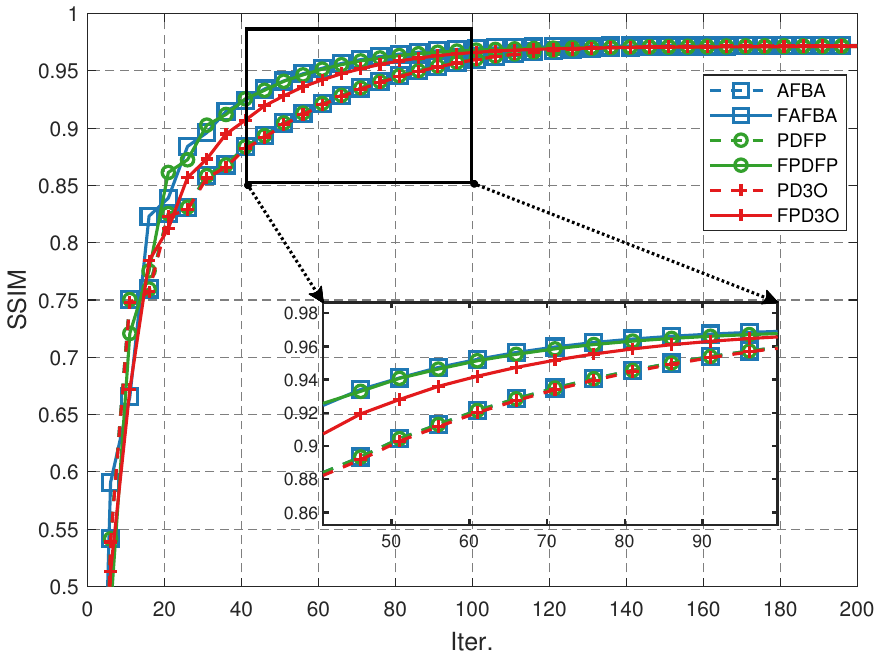}
	\caption{Evolutions of the SNR values and SSIM values with respect to the numbers of iteration.}
	\label{Fig:CTV_SNR_SSIM}
\end{figure*}

\begin{figure}[htb]
	\centering
	\subfloat{\includegraphics[width=0.3\linewidth]{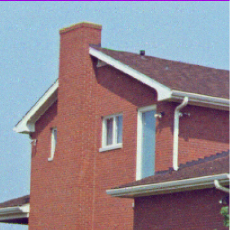}}
	\subfloat{\includegraphics[width=0.3\linewidth]{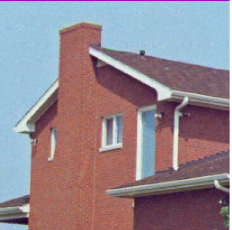}}
	\subfloat{\includegraphics[width=0.3\linewidth]{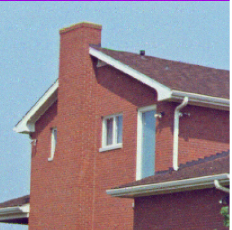}}\\
	\vspace{-1em}
	\subfloat{\includegraphics[width=0.3\linewidth]{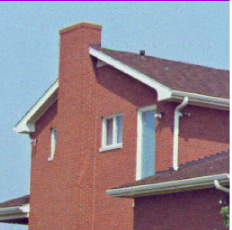}}
	\subfloat{\includegraphics[width=0.3\linewidth]{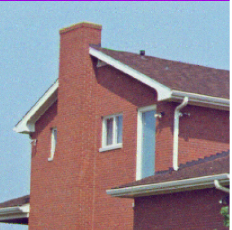}}
	\subfloat{\includegraphics[width=0.3\linewidth]{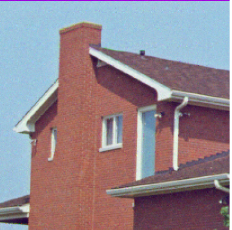}}
	\caption{Restored images by AFBA, PDFP, and PD3O (first line); FAFBA, FPDFP, and FPD3O (second line), respectively.}
	\label{Fig:CTV_re}
\end{figure}

In order to compare the differences between the primal dual algorithms more graphically, we plot the evolution of the SNR values and SSIM values with respect to the numbers of iterations for all tested algorithms in Figure \ref{Fig:CTV_SNR_SSIM}. We can see from Figure \ref{Fig:CTV_SNR_SSIM} that, the fair primal dual methods  
outperform the original primal dual methods before all tested primal dual methods converge to a restored image with almost the same quality. Moreover, we present the images restored by these primal dual methods in Figure \ref{Fig:CTV_re}.

\section{Conclusion}\label{Conc}

In this study, we tackled the image inverse problems prevalent in fields such as image processing, medical imaging, and computer vision, focusing on the mathematical challenge of reconstructing unknown images from observed measurements. We introduced a novel Fair Primal Dual (FPD) algorithmic framework that balances the computational demands between primal and dual subproblems by distributing the data-fidelity term into both subproblems. This approach addresses the traditional skew in such algorithms that often favor one subproblem over the other, thereby enhancing both convergence and effectiveness. To overcome the challenges posed by the new dual subproblems, we developed an Inexact Fair Primal Dual (IFPD) framework that allows for approximate solutions, ensuring practical applicability without sacrificing robustness. Our methods demonstrated superior performance in image denoising and super-resolution tasks, showing significant improvements over the original primal dual framework. In the future, we aim to design accelerated versions of the FPD algorithmic framework to develop more efficient algorithms for image inverse problems.

%\section*{Acknowledgments}
%This should be a simple paragraph before the References to thank those individuals and institutions who have supported your work on this article.

{\appendices
	\section{Proof of Theorem \ref{Thm1:FPD}}\label{APP:thm1}
	\begin{IEEEproof}
	We consider the four fair primal dual algorithms, respectively.
	
	\textbf{Case: Condat-V$\mathbf{\tilde{u}}$.}
	From the optimality condition of \eqref{FPDx}, we have that for any $x$,
	\begin{align}
		&h(x)-h(\hat{x}^{k+1})+\langle \nabla f_1(x^k)+y^k,x-\hat{x}^{k+1}\rangle\notag\\
		\ge&\langle\frac{1}{\sigma}(x^{k}-\hat{x}^{k+1}),x-\hat{x}^{k+1}\rangle.\label{conv1_1}
	\end{align}
	Since $f_1$ is convex and $\nabla h_1$ is $L_{f_1}$-Lipschitz continuous, from \eqref{Lsmooth1} we have that for any $x$
	\begin{align}
		&f_1(x)-f_1(\hat{x}^{k+1})-\langle \nabla f_1(x^k),x-\hat{x}^{k+1}\rangle \notag\\
		\ge&-\frac{L_{f_1}}{2}\|\hat{x}^{k+1}-x^k\|^2. \label{conv1_2}
	\end{align}
	Combining \eqref{conv1_1} and \eqref{conv1_2} yields that for any $x$,
	\begin{align}
		&h(x)+f_1(x)-h(\hat{x}^{k+1})-f_1(\hat{x}^{k+1})+\langle x-\hat{x}^{k+1},y^{k+1}\rangle \notag\\
		\ge& \langle x-\hat{x}^{k+1},(y^{k+1}-y^{k})+\frac{1}{\sigma}(x^k-\hat{x}^{k+1})\rangle\label{conv1_3}\\
		&+\frac{L_{f_1}}{2}\|\hat{x}^{k+1}-x^k\|^2.\notag
	\end{align}
	Similarly from the optimality of \eqref{FPDy} and Iteration I' of Condat-V\~{u}, we obtain that for any $y$,
	\begin{align}
		&(\tilde{g}+f_2)^*(y)-(\tilde{g}+f_2)^*(y^{k+1})-\langle y-y^{k+1},\hat{x}^{k+1}\rangle \notag \\
		\ge&\langle y-y^{k+1}, \frac{1}{\tau}({y^k-y^{k+1}})+(\hat{x}^{k+1}-x^k)\rangle.\label{conv1_4}
	\end{align}
	Summing \eqref{conv1_3} and \eqref{conv1_4} and rearranging terms, we obtain that for any $\mathbf{v}$,
	\begin{align}\label{conv1_5}
		&\Psi(\mathbf{u}^{k+1})-\Psi(\mathbf{v})-\langle \mathbf{v}-\mathbf{u}^{k+1}, F(\mathbf{u}^{k+1})\rangle \notag\\
		\le&\langle \mathbf{v}-\mathbf{v}^{k+1},Q(\mathbf{v}^{k+1}-\mathbf{v}^k)\rangle +\frac{L_{h_1}}{2}\|x^k-x^{k+1}\|^2 \notag\\
		=&\frac{1}{2}\left(\|\mathbf{v}-\mathbf{v}^{k}\|_Q^2-\|\mathbf{v}-\mathbf{v}^{k+1}\|_Q^2-\|\mathbf{v}^{k+1}-\mathbf{v}^{k}\|_{M}^2\right).
	\end{align}
	where $\mathbf{u}^{k+1}=\begin{pmatrix} \hat{x}^{k+1}\\ y^{k+1}\end{pmatrix}$, $Q = \begin{pmatrix}
		\frac{1}{\sigma}I & -I\\-I & \frac{1}{\tau}I
	\end{pmatrix}$, $M:=Q-\begin{pmatrix}
		L_{f_1}I & 0\\0 & 0
	\end{pmatrix}$, and the equality is due to Lemma \ref{ident2}. Hence, we can obtain \eqref{Thm1:key} by setting $d_{\text{CV}}(\mathbf{v};\mathbf{v^k}):=\frac{1}{2}\|\mathbf{v}-\mathbf{v}^{k}\|_Q^2$ and $\tilde{d}_{\text{CV}}(\mathbf{v}^{k+1};\mathbf{v}^k):=\frac{1}{2}\|\mathbf{v}^{k+1}-\mathbf{v}^{k}\|_M^2$.	
	
	\textbf{Case: PDFP.} Similar to \eqref{conv1_1}, from the Iteration II' of PDFP, we have that for any $x$,
	\begin{align}
		&h(x)-h({x}^{k+1})+\langle \nabla f_1(x^k)+y^k,x-{x}^{k+1}\rangle\notag\\
		\ge&\langle\frac{1}{\sigma}(x^{k}-{x}^{k+1}),x-{x}^{k+1}\rangle.\label{conv1_6}
	\end{align}
	Setting $x=x^{k+1}$ in \eqref{conv1_1} and adding it with \eqref{conv1_6} yields,
	\begin{align}
		&h(x)+f_1(x)-h(\hat{x}^{k+1})-f_1(\hat{x}^{k+1})+\langle x-\hat{x}^{k+1},y^{k+1}\rangle \notag\\
		\ge&\langle x-x^{k+1}, \frac{1}{\sigma}(\hat{x}^{k+1}-x^{k+1})\rangle+\langle x-\hat{x}^{k+1}, \frac{1}{\sigma}({x}^{k}-\hat{x}^{k+1})\rangle \notag\\
		&+\langle x^{k+1}-\hat{x^{k+1}},y^{k+1}-y^k\rangle+\frac{1}{2}L_{f_1}\|\hat{x}^{k+1}-x^k\|^2,\label{conv1_7}
	\end{align}
	where the inequality is due to \eqref{conv1_2}.
	From the optimality of \eqref{FPDy} and Iteration I' of PDFP, we obtain that for any $y$,
	\begin{align}
		&(\tilde{g}+f_2)^*(y)-(\tilde{g}+f_2)^*(y^{k+1})-\langle y-y^{k+1},\hat{x}^{k+1}\rangle \notag \\
		\ge&\langle y-y^{k+1}, \frac{1}{\tau}({y^k-y^{k+1}})\rangle.\label{conv1_8}
	\end{align}
	Summing \eqref{conv1_7} and \eqref{conv1_8} and rearranging terms, we obtain that for any $\mathbf{v}$,
	\begin{align*}%\label{conv1_9}
		&\Psi(\mathbf{u}^{k+1})-\Psi(\mathbf{v})-\langle \mathbf{v}-\mathbf{u}^{k+1}, F(\mathbf{u}^{k+1})\rangle \notag\\
		\le& \frac{1}{2}\left(\|\mathbf{v}-\mathbf{v}^k\|^2_Q-\|\mathbf{v}-\mathbf{v}^{k+1}\|^2_Q\right)\\
		&-\frac{1}{2}\left(\|x^{k+1}-x^k\|^2_{(\frac{1}{\sigma }-L_{f_1})I}+\|y^{k+1}-y^k\|^2_{(\frac{1}{\tau}-\sigma)I}\right),
	\end{align*}
	where $Q=\begin{pmatrix}
		\frac{1}{\sigma}I & 0\\ 0 & \frac{1}{\tau}I
	\end{pmatrix}$. Hence we can obtain \eqref{Thm1:key} of PDFP by setting $d_{\text{PDFP}}(\mathbf{v};\mathbf{v}^{k}):=\frac{1}{2}\|\mathbf{v}-\mathbf{v}^{k}\|^2_Q$ and $\tilde{d}_{\text{PDFP}}(\mathbf{v^{k+1}};\mathbf{v}^{k}):=\frac{1}{2}\|\mathbf{v}^{k+1}-\mathbf{v}^{k}\|^2_M$, where $M:=\begin{pmatrix}
		(\frac{1}{\sigma }-L_{f_1})I & 0 \\ 0 & (\frac{1}{\tau}-\sigma)I
	\end{pmatrix}$.

	\textbf{Case: AFBA.} Summing \eqref{conv1_3} and \eqref{conv1_8} yields that
	\begin{align}\label{conv1_10}
		&\Psi(\mathbf{u}^{k+1})-\Psi(\mathbf{v})-\langle \mathbf{v}-\mathbf{u}^{k+1}, F(\mathbf{u}^{k+1})\rangle \notag\\
		\le& \frac{1}{2\sigma}\left(\|x-x^k\|^2-\|x-\hat{x}^{k+1}\|^2-\|\hat{x}^{k+1}-x^k\|^2\right) \notag\\
		&-\langle x-\hat{x}^{k+1},y^{k+1}-y^k\rangle+\frac{L_{f_1}}{2}\|\hat{x}^{k+1}-x^k\|^2 \\
		&+\frac{1}{2\tau}\left(\|y-y^k\|^2-\|y-y^{k+1}\|^2-\|y^{k+1}-y^k\|^2\right),\notag
	\end{align}
	From the Iteration II' of AFBA, we have that
	\begin{align}
		&\frac{1}{2\sigma}\|x-\hat{x}^{k+1}\|^2+\langle x-\hat{x}^{k+1},y^{k+1}-y^k\rangle \notag\\
		=&\frac{1}{2\sigma}\|x-{x}^{k+1}\|^2-\frac{\sigma}{2}\|y^{k+1}-y^k\|^2. \label{conv1_11}
	\end{align}
	Substituting \eqref{conv1_11} into \eqref{conv1_10} yields that, for any $\mathbf{v}$,
	\begin{align}
		&\Psi(\mathbf{u}^{k+1})-\Psi(\mathbf{v})-\langle v-\mathbf{u}^{k+1}, F(\mathbf{u}^{k+1})\rangle \notag\\
		\le &\frac{1}{2}\left(\|\mathbf{v}-\mathbf{v}^{k}\|_Q^2-\|\mathbf{v}-\mathbf{v}^{k+1}\|_Q^2-\|\mathbf{v}^{k+1}-\mathbf{v}^{k}\|_{M}^2\right),
	\end{align}
	where $Q$ and $M$ are the same as PDFP's. Therefore, we directly set $d_{\text{AFBA}}(\mathbf{v};\mathbf{v}^{k}):=d_{\text{PDFP}}(\mathbf{v};\mathbf{v}^{k})$ and $\tilde{d}_{\text{AFBA}}(\mathbf{v}^{k+1};\mathbf{v}^{k}):=\tilde{d}_{\text{AFBA}}(\mathbf{v}^{k+1};\mathbf{v}^{k})$ to obtain \eqref{Thm1:key}.
	
	\textbf{Case: PD3O.} We firstly define 
	\begin{align}
		&\bar{d}_{\text{PD3O}}(\mathbf{v},w;\mathbf{v}',w') \notag\\
		&:=d_{\text{PDFP}}(\mathbf{v};\mathbf{v}')+\frac{\sigma}{2}\|\nabla f_{1}(w)-\nabla f_{1}(w')\|^2 \notag\\
		&\quad -\langle y-y',(x-x')-\sigma(\nabla f_{1}(w)-\nabla f_{1}(w'))\rangle \notag\\
		&\quad -\langle x-x',\nabla f_{1}(w)-\nabla f_{1}(w')\rangle.
	\end{align}
	From the Iteration I' of PD3O and the optimality conditions of \eqref{FPDx} and \eqref{FPDy}, we have that for any $\mathbf{v}$,
	\begin{align}
		&\Psi(\mathbf{u}^{k+1})-\Psi(\mathbf{v})-\langle v-\mathbf{u}^{k+1}, F(\mathbf{u}^{k+1})\rangle \notag\\
		\le&{d}_{\text{PDFP}}(\mathbf{v};\mathbf{v}^k)-d_{\text{PDFP}}(\mathbf{v};\mathbf{v}^{k+1})-d_{\text{PDFP}}(\mathbf{v}^{k+1};\mathbf{v}^k)\notag\\
		&+\langle x-x^{k+1}, y^{k+1}-y^k\rangle+\langle y-y^{k+1},x^{k+1}-x^k\rangle\notag\\
		&-\langle y-y^{k+1}, \sigma(\nabla f_{1}(x^{k+1})-\nabla f_{1}(x^k))\rangle \notag\\
		&+f_1({x}^{k+1})-f_1(x)+\langle \nabla f_1(x^k),x-x^{k+1}\rangle \notag\\
		=&\bar{d}_{\text{PD3O}}(\mathbf{v},\nabla f_1(x);\mathbf{v}^k,\nabla f_1(x^k)) \notag\\
		&-\bar{d}_{\text{PD3O}}(\mathbf{v},\nabla f_1(x);\mathbf{v}^{k+1},\nabla f_1(x^{k+1}))\notag\\
		&-\bar{d}_{\text{PD3O}}(\mathbf{v}^{k+1},\nabla f_1(x);\mathbf{v}^k,\nabla f_1(x^k)) \notag\\
		&+f_1({x}^{k+1})-f_1(x)+\langle \nabla f_1(x^{k+1}),x-x^{k+1} \notag \rangle\\
		&+\frac{\sigma}{2}\|\nabla f_{1}(x)-\nabla f_{1}(x^{k+1})\|^2 \notag\\
		\le&d_{\text{PD3O}}(\mathbf{v};\mathbf{v}^k)-d_{\text{PD3O}}(\mathbf{v};\mathbf{v}^{k+1})-\tilde{d}_{\text{PD3O}}(\mathbf{v}^{k+1};\mathbf{v}^k),
	\end{align}
	where the second inequality is due to \eqref{Lsmooth2}, 
	$d_{\text{PD3O}}(\mathbf{v};\mathbf{v}^k):=\bar{d}_{\text{PD3O}}(\mathbf{v},\nabla f_1(x);\mathbf{v}^k,\nabla f_1(x^k))$ and
	\begin{align}
		&\tilde{d}_{\text{PD3O}}(\mathbf{v}^{k+1};\mathbf{v}^k):=\bar{d}_{\text{PD3O}}(\mathbf{v}^{k+1},\nabla f_1(x);\mathbf{v}^k,\nabla f_1(x^k))\notag \\ &\quad\quad\quad\quad+\left(\frac{1}{2L_{f_1}}-\frac{\sigma}{2}\right)\|\nabla f_{1}(x)-\nabla f_{1}(x^{k+1})\|^2.
	\end{align}
\end{IEEEproof}

	\section{Proof of Theorem \ref{Thm2:FPD}}\label{App:thm2}
	\begin{IEEEproof}
		We consider the four fair primal dual algorithms, respectively.
	
	\textbf{Case: Condat-V$\mathbf{\tilde{u}}$.} If $\sigma\tau<1-\sigma L_{f_1}$, $Q, M$ defined for Condat-V\~{u} algorithm in Theorem \ref{Thm1:FPD} are positive definite and $d_{\text{CV}}(\mathbf{v};\mathbf{v}^k)>0, \tilde{d}_{\text{CV}}(\mathbf{v}^{k+1};\mathbf{v}^k)>0$, accordingly.
	
	\textbf{Case: PDFP.} It can be seen that matrix $Q$ defined for PDFP algorithm in Theorem \ref{Thm1:FPD} is positive definite and $M$ is also positive definite owing to $\sigma\tau<1$ and $\sigma L_{f_1}<1$ in Table \ref{Tab:condition}. Hence $d_{\text{PDFP}}(\mathbf{v};\mathbf{v}^k)>0$, and $\tilde{d}_{\text{PDFP}}(\mathbf{v}^{k+1};\mathbf{v}^k)>0$.
	
	\textbf{Case: AFBA.} Since $d_{\text{AFBA}}(\mathbf{v};\mathbf{v}^{k}):=d_{\text{PDFP}}(\mathbf{v};\mathbf{v}^{k})$ and $\tilde{d}_{\text{AFBA}}(\mathbf{v}^{k+1};\mathbf{v}^{k}):=\tilde{d}_{\text{AFBA}}(\mathbf{v}^{k+1};\mathbf{v}^{k})$, wo can prove $\tilde{d}_{\text{PDFP}}(\mathbf{v}^{k+1};\mathbf{v}^k)>0$ with $\sigma\tau<1$ and $\sigma L_{f_1}<1$.
	
	\textbf{Case: PD3O.} From the definition of $d_{\text{AFBA}}(\mathbf{v};\mathbf{v}^{k})$, we have that for any $\mathbf{v},\mathbf{v}^{k}$
	\begin{align}
		&\quad d_{\text{PD3O}}(\mathbf{v};\mathbf{v}^{k}) \notag\\
		&> \frac{1}{2\sigma}\|x-x^k\|^2+\frac{\sigma}{2}\|y-y^k\|^2+\frac{\sigma}{2}\|\nabla f_{1}(x)-\nabla f_{1}(x^k)\|^2 \notag\\
		&\quad-\langle y-y^k,(x-x^k)-\sigma(\nabla f_{1}(x)-\nabla f_{1}(x^k))\rangle \notag\\
		&\quad-\langle x-x^k,\nabla f_{1}(x)-\nabla f_{1}(x^k)\rangle \notag\\
		&=\frac{1}{2}\left\|\frac{1}{\sqrt{\sigma}}(x-x^k)-\sqrt{\sigma}\left(y-y^k+\nabla f_{1}(x)-\nabla f_{1}(x^k)\right)\right\|^2\notag\\
		&\ge0,
	\end{align}
	where the inequality is due to $\sigma\tau<1$. Then it follows from $\sigma L_{f_1}<1$ that
	$\tilde{d}_{\text{PD3O}}(\mathbf{v}^{k+1};\mathbf{v}^{k})>0$ for $k\ge0$.
	\end{IEEEproof}
	
\section{Proof of Theorem \ref{Thm2+:FPD}}\label{App:thm2+}
	
	Summing \eqref{Thm1:key} from $k=0$ to $N-1$ yields that
\begin{align}\label{pro1_1}
	&\sum_{k=0}^{N-1}\left(\Phi(\mathbf{u}^{k+1})-\Phi(\mathbf{v})-\langle \mathbf{v}-\mathbf{u}^{k+1},F(\mathbf{u}^{k+1})\rangle\right)\\
	&\le d(\mathbf{v};\mathbf{v}^{0})-d(\mathbf{v};\mathbf{v}^{N})-\sum_{k=0}^{N-1}\tilde{d}(\mathbf{v}^{k+1};\mathbf{v}^{k}).\notag%-\frac{1}{2}\|v-v^{K}\|^2_H.
\end{align}
For any $\mathbf{v}$, $\Phi(\mathbf{v})-\Phi(\mathbf{v}^*)-\langle \mathbf{v}^*-\mathbf{v},F(\mathbf{v}^*)\rangle\ge0$ holds for any saddle point $\mathbf{v}^*$ as defined in \eqref{saddlePD}. Hence, combining \eqref{pro1_1} with $\mathbf{v}=\mathbf{v}^*$, we conclude that $\{d(\mathbf{v}^*;\mathbf{v}^{k})\}$ is bounded and $\{\tilde{d}(\mathbf{v}^{k+1};\mathbf{v}^{k})\}$ is summable. 
For Condat-V\~u, PDFP, and AFBA algorithms, from the positive definiteness of $Q$ and $\bar{Q}$ we can conclude that  $\{\mathbf{v}^k\}$ is bounded and $\|\mathbf{v}^{k+1}-\mathbf{v}^k\|\to0$ as $k\to\infty$. For PD3O algorithm, from the definition of $\tilde{d}_{\text{PD3O}}(\mathbf{v}^{k+1};\mathbf{v}^k)$ and convergence condition, we have $\nabla f_1(x^*)-\nabla f_1(x^{k+1})\to 0$ as $k\to 0$. Then
$\tilde{d}_{\text{PD3O}}(\mathbf{v}^{k+1};\mathbf{v}^k)=\frac{1}{2}\|\mathbf{v}^{k+1}-\mathbf{v}^k\|^2_M\to 0,$ as $k\to 0$, where 
$M=\begin{pmatrix}
    \frac{1}{\sigma}I & -I\\-I & \frac{1}{\tau}I
\end{pmatrix}$
is a positive matrix. Hence $\|\mathbf{v}^{k+1}-\mathbf{v}^k\|\to0$ as $k\to\infty$ for PD3O algorithm.
We can also prove $\hat{x}^k-x^k\to 0$ as $k\to\infty$ from Iteration I' and II'.

Based on above analysis, $\{\mathbf{v}^k\}$ possesses at least one cluster point $\mathbf{v}^{\infty}$ and there exists a subsequence $\{\mathbf{v}^{k_j}\}$ such that $\mathbf{v}^{k_j}\to \mathbf{v}^{\infty}$ as $j\to\infty$. Taking $j\to\infty$, we have
\begin{align*}
	&f_1(x)+h(x)-f(\hat{x}^{k_j+1})-h(\hat{x}^{k_j+1})+\langle x-\hat{x}^{k_j+1},{y}^{k_j+1}\rangle\notag\\
	&\ge\langle x-\hat{x}^{k_j+1},({y}^{k_j+1}-y^{k_j})-\frac{1}{\sigma}(\hat{x}^{k_j+1}-x^{k_j})\rangle\\
	&\quad-\frac{L_{f_1}}{2}\|\hat{x}^{k_j+1}-x^{k_j}\|^2.
\end{align*}
For Condat-V\~u algorithm, 
\begin{align*}
	&(\tilde{g}+f_2)^*(y)-(\tilde{g}+f_2)^*({y}^{k_j+1})+\langle y-{y}^{k_j+1},- {x}^{k_j+1}\rangle\\
	\ge& \langle y-{y}^{k_j+1}, ({x}^{k_j+1}-x^{k_j})-\frac{1}{\tau }({y}^{k_j+1}-y^{k_j})\rangle.
\end{align*}
For PDFP and AFBA algorithms,
\begin{align*}
    &(\tilde{g}+f_2)^*(y)-(\tilde{g}+f_2)^*(y^{k_j+1})-\langle y-y^{k_j+1},\hat{x}^{k_j+1}\rangle \notag \\
    \ge&\langle y-y^{k_j+1}, \frac{1}{\tau}({y^{k_j}-y^{k_j+1}})\rangle.
\end{align*}
For PD3O algorithm,
\begin{align*}
    &(\tilde{g}+f_2)^*(y)-(\tilde{g}+f_2)^*(y^{k_j+1})-\langle y-y^{k_j+1},\hat{x}^{k_j+1}\rangle \notag \\
    \ge&\langle y-y^{k_j+1}, \frac{1}{\tau}({y^{k_j}-y^{k_j+1}})+(\hat{x}^{k_j+1}-x^{k_j})\\
    & +\sigma(\nabla f_1(x^{k_j})-\nabla f_1(\hat{x}^{k_j+1}))\rangle.
\end{align*}
Passing $j\to \infty$, we conclude that $\mathbf{v}^{\infty}$ is a saddle point of \eqref{saddle2 pro}. On the other hand, by summing the inequality \eqref{Thm1:key} over $k_j,k_j+1,\cdots,N-1$ and taking $\mathbf{v}=\mathbf{v}^{\infty}$, we obtain that for any
$N\ge k_j+1$,
\begin{align*}
	d(\mathbf{v}^N;\mathbf{v}^{\infty})\le d(\mathbf{v}^{k_j};\mathbf{v}^{\infty}).
\end{align*}
Since the right side of the above inequality tends to 0 as $j\to\infty$, this implies that $\mathbf{v}^N\to \mathbf{v}^{\infty}$ as $N\to\infty$. We also conclude that $\mathbf{v}^k\to \mathbf{v}^{\infty}$ since $\|\mathbf{v}^k-\mathbf{v}^{k+1}\|\to0$ as $k\to\infty$.
	
\section{Proof of Theorem \ref{Thm3:FPD}}\label{App:thm3}
\begin{IEEEproof}
	Summing \eqref{Thm1:key} from $k=0$ to $N-1$ yields that
	\begin{align}\label{APP3_1}
		&\sum_{k=0}^{N-1}\left(\Psi(\mathbf{u}^{k+1})-\Phi(\mathbf{v})-\langle \mathbf{v}-\mathbf{u}^{k+1},F(\mathbf{u}^{k+1})\rangle\right) \notag\\
		&\le d(\mathbf{v};\mathbf{v}^{0})-d(\mathbf{v};\mathbf{v}^{N})-\sum_{k=0}^{N-1}\tilde{d}(\mathbf{v}^{k+1};\mathbf{v}^{k})\notag\\
		&\le d(\mathbf{v};\mathbf{v}^{0}) \le c_0\|\mathbf{v}-\mathbf{v}^0\|^2,
	\end{align}
	where $c_0>0$ is a constant and the second inequality is due to Theorem \ref{Thm2:FPD}. From the definition of $d(\mathbf{v};\mathbf{v^k})$, the last inequality is direct for Condat-V\~{u}, PDFP, and AFBA algorithms. And the last inequality also holds for the PD3O algorithm by exploiting Cauchy Schwarz inequality and L-smoothness of $f_1$ into $d_{\text{PD3O}}(\mathbf{v};\mathbf{v^k})$.
	Since $\psi(\cdot)$ is convex, we have that
	\begin{align*}
		&\Psi(\mathbf{\hat{u}}^N)-\Psi(\mathbf{v})-\langle \mathbf{v}-\mathbf{\hat{u}}^N,F(\mathbf{v})\rangle\notag\\
		\le&\frac{1}{N}\sum_{k=0}^{N-1}\left(\Psi(\mathbf{u}^{k+1})-\Phi(\mathbf{v})-\langle \mathbf{v}-\mathbf{u}^{k+1},F(\mathbf{u}^{k+1})\rangle\right).
	\end{align*}
	So we obtain the ergodic convergence rate by setting $\mathbf{v}=\mathbf{v}^*$ in \eqref{APP3_1}.
	%And the pointwise rate of convergence, in terms of the residual between two successive iterates, can be obtained by following the analysis in \cite{HY15}. So we omit here.
\end{IEEEproof}
\section{Proof of Theorem 
\ref{Thm6:IFPD}}\label{App:thm6}
\begin{IEEEproof}
From the error criterion \eqref{IFPDd}, we get that
\begin{equation*}
	\|d^{k+1}\|\|y^{k+1}\|\le\epsilon_{k+1},\|d^{k+1}\|\le\epsilon_{k+1}.
\end{equation*}
Summing \eqref{Thm4:key} from $k=0$ to $N-1$ yields that
\begin{align*}
	&\sum_{k=0}^{N-1}\left(\Psi(\mathbf{u}^{k+1})-\Phi(\mathbf{v})-\langle \mathbf{v}-\mathbf{u}^{k+1},F(\mathbf{u}^{k+1})\rangle\right) \notag\\
	\le&d(\mathbf{v};\mathbf{v}^{0})-d(\mathbf{v};\mathbf{v}^{N})-\sum_{k=0}^{N-1}\tilde{d}(\mathbf{v}^{k+1};\mathbf{v}^{k})\notag\\
	&+\sum_{k=0}^{N-1}\frac{1}{\tau}\langle y-y^{k+1},d^{k+1}\rangle\notag\\
	\le&d(\mathbf{v};\mathbf{v}^{0})-d(\mathbf{v};\mathbf{v}^{N})-\sum_{k=0}^{N-1}\tilde{d}(\mathbf{v}^{k+1};\mathbf{v}^{k})\notag\\
        &+\frac{1}{\tau}(1+\|y\|)\sum_{k=0}^{N-1}\epsilon_{k+1}.
\end{align*}
Since $\{\epsilon_{k}\}$ is summable, we can prove the convergence and convergence rate of Algorithm \ref{Alg:IFPD} by following the proof of Theorem \ref{Thm2+:FPD} and Theorem \ref{Thm3:FPD}, respectively.	
\end{IEEEproof}

%	\section{Proof of Theorem 
%	\ref{Thm6:IFPD}}\label{App:thm6}
%	Similar as the proof of Theorem \ref{Thm6:IFPD}.
}

%Use $\backslash${\tt{appendix}} if you have a single appendix:
%Do not use $\backslash${\tt{section}} anymore after $\backslash${\tt{appendix}}, only $\backslash${\tt{section*}}.
%If you have multiple appendixes use $\backslash${\tt{appendices}} then use $\backslash${\tt{section}} to start each appendix.
%You must declare a $\backslash${\tt{section}} before using any $\backslash${\tt{subsection}} or using $\backslash${\tt{label}} ($\backslash${\tt{appendices}} by itself
% starts a section numbered zero.)}

%{\appendices
%\section*{Proof of the First Zonklar Equation}
%Appendix one text goes here.
% You can choose not to have a title for an appendix if you want by leaving the argument blank
%\section*{Proof of the Second Zonklar Equation}
%Appendix two text goes here.}

\bibliographystyle{IEEEtran}
\bibliography{IEEEabrv,QuArt,QuBook}

\newpage

%\section{Biography Section}

%If you have an EPS/PDF photo (graphicx package needed), extra braces are
% needed around the contents of the optional argument to biography to prevent
% the LaTeX parser from getting confused when it sees the complicated
% $\backslash${\tt{includegraphics}} command within an optional argument. (You can create
% your own custom macro containing the $\backslash${\tt{includegraphics}} command to make things
% simpler here.)
%\vspace{11pt}
%\bf{If you include a photo:}\vspace{-33pt}

\begin{IEEEbiography}[{\includegraphics[width=1in,height=1.25in,clip,keepaspectratio]{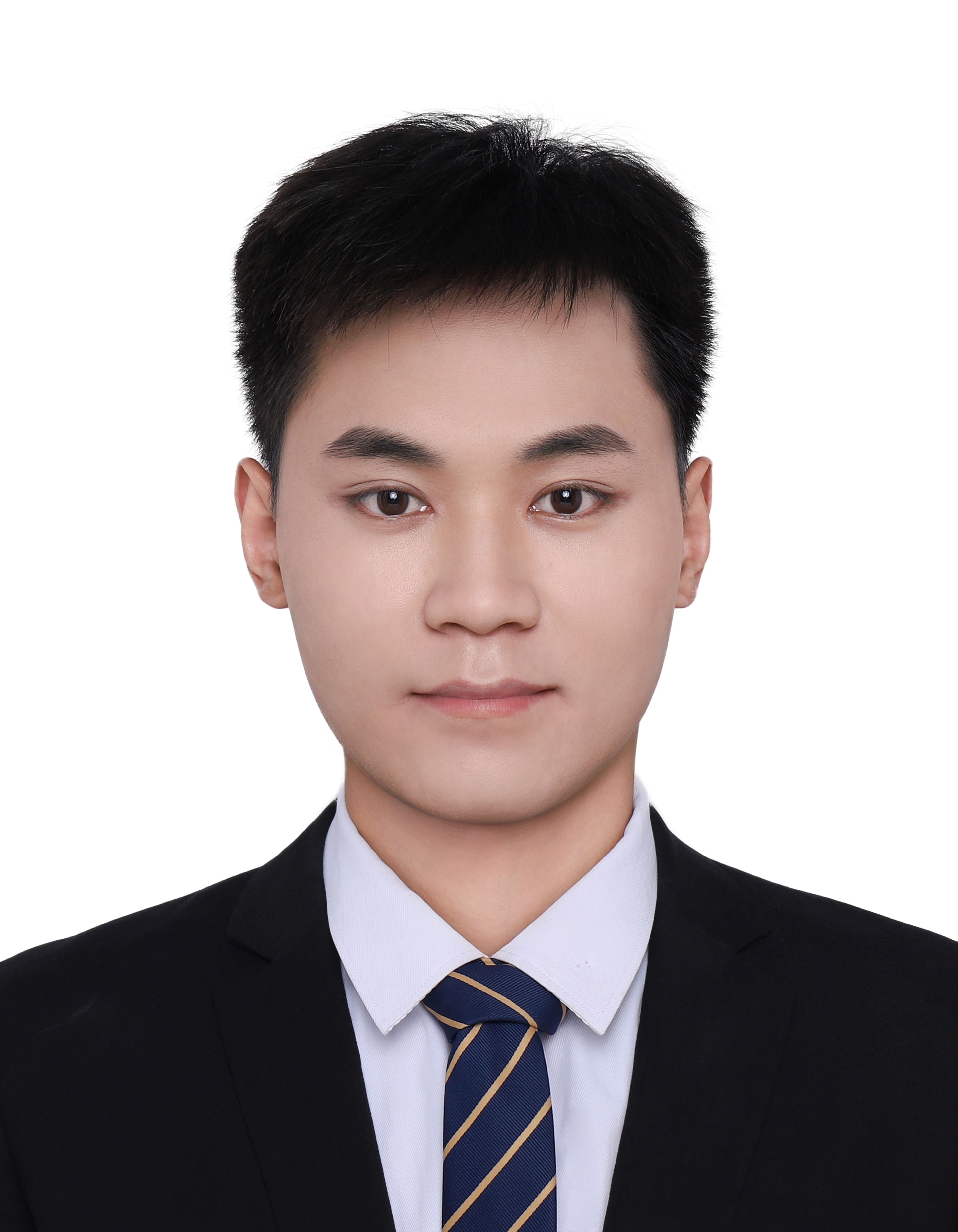}}]{Yunfei Qu}
received the B.S. degree in Mathematics and Applied Mathematics from Hangzhou Dianzi University, Hangzhou, China, in 2019. He is currently pursuing the Ph.D. degree in LMIB of the Ministry of Education, School of Mathematical Sciences, Beihang University, Beijing, China. His current research focuses on first-order optimization algorithms, splitting methods, and their applications.
\end{IEEEbiography}
\begin{IEEEbiography}[{\includegraphics[width=1in,height=1.25in,clip,keepaspectratio]{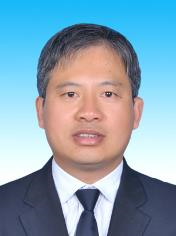}}]{Deren Han}
	is currently a professor, doctoral supervisor, and the dean of the School of Mathematical Sciences, Beihang University. 
	He received the B.S. and Ph.D. degrees at Nanjing University, China, in 1997 and 2002, respectively.
	
	His research mainly focuses on numerical methods for large-scale optimization problems and variational inequality problems, as well as their applications in transportation planning and magnetic resonance imaging.
	He has published more than 150 papers in journals and conferences, 
	and won the Youth Operations Research Award of Operations Research Society of China, the Second prize of Science and Technology Progress of Jiangsu Province and other awards. He has presided over the National Science Fund for Distinguished Young Scholars and other projects.
	Professor Deren Han also serves as the editorial board member of Numerical Computation and Computer Applications, Journal of the Operations Research Society of China, Journal of Global Optimization, and Asia-Pacific Journal of Operational Research.
\end{IEEEbiography}

%\vspace{11pt}
%
%\bf{If you will not include a photo:}\vspace{-33pt}
%\begin{IEEEbiographynophoto}{John Doe}
%Use $\backslash${\tt{begin\{IEEEbiographynophoto\}}} and the author name as the argument followed by the biography text.
%\end{IEEEbiographynophoto}

\vfill

\end{document}